\documentclass[lettersize,journal]{IEEEtran}
\usepackage{amsmath,amsfonts}
\usepackage{amssymb}
\usepackage{algorithmic}
\usepackage{algorithm}
\usepackage{array}
\usepackage[caption=false,font=normalsize,labelfont=sf,textfont=sf]{subfig}
\usepackage{textcomp}
\usepackage{stfloats}
\usepackage{url}
\usepackage{makecell}
\usepackage{verbatim}
\usepackage{graphicx}
\usepackage{cite}
\usepackage[utf8]{inputenc} 
\usepackage[T1]{fontenc}    

\usepackage{booktabs}       
\usepackage{nicefrac}       
\usepackage{microtype}      
\usepackage{multicol}
\usepackage{color}
\usepackage[table]{xcolor}

\usepackage{bm}
\usepackage{cuted}
\usepackage{mathrsfs}
\usepackage{mathtools}
\usepackage{multirow}
\usepackage{makecell}
\usepackage{subcaption}
\usepackage{enumerate}
\usepackage{float}				  
\usepackage{enumitem}

\begin{document}

\title{ConvD: Attention Enhanced Dynamic Convolutional Embeddings for Knowledge Graph Completion}

\author{Wenbin Guo, Zhao Li,
        Xin Wang,~\IEEEmembership{Member,~IEEE,}
        Zirui Chen,
        Jun Zhao, \ \ \ \ \  \ \ \ \ \ \ \ \ \ \ \ \ \ \ \ \ \ \ \ \ \ \ \ \ \ \ \ \ \ \ \  \ \ \ \ \ \ \ \ \ \ \ \ \ \ \ \ \ 
        Jianxin Li,~\IEEEmembership{Senior Member,~IEEE,}
        Ye Yuan,~\IEEEmembership{Member,~IEEE}
        
\thanks{Manuscript accepted June, 2025. (Corresponding author: Xin Wang.)}
\thanks{(Wenbin Guo and Zhao Li contributed equally to this work.)}

\thanks{This work is supported by the National Natural Science Foundation of China (62472311), the Key Research and Development Program of Ningxia Hui Autonomous Region (2023BEG02067), and the Australia research council discovery project (DP240101591).}

\IEEEcompsocitemizethanks{\IEEEcompsocthanksitem Wenbin Guo, Zhao Li, Xin Wang, and Zirui Chen are with the College of Intelligence and Computing, Tianjin University, Tianjin 300354, China, and also with the Tianjin Key Laboratory of Cognitive Computing and Application, Tianjin 300354, China. E-mail: \{Wenff, lizh, wangx, zrchen\}@tju.edu.cn
\IEEEcompsocthanksitem Jun Zhao is with the School of Economics and Management, Ningxia University, Yinchuan 750021, China. E-mail: tzhaoj@nxu.edu.cn
\IEEEcompsocthanksitem Jianxin Li is with the Discipline of Busin
ess Systems and Operations, Edith Cowan University, Joondalup WA 6027, Australia. E-mail: jianxin.li@deakin.edu.au
\IEEEcompsocthanksitem Ye Yuan is with the School of Computer Science and Technology, Beijing Institute of Technology, Beijing 100081, China. E-mail: yuan-ye@bit.edu.cn}

}

\markboth{Journal of \LaTeX\ Class Files,~Vol.~14, No.~8, August~2024}%
{Shell \MakeLowercase{\textit{et al.}}: A Sample Article Using IEEEtran.cls for IEEE Journals}

\maketitle

\begin{abstract}
Knowledge graphs often suffer from incompleteness issues, which can be alleviated through information completion. However, current state-of-the-art deep knowledge convolutional embedding models rely on external convolution kernels and conventional convolution processes, which limits the feature interaction capability of the model. This paper introduces a novel dynamic convolutional embedding model, ConvD, which directly reshapes relation embeddings into multiple internal convolution kernels. This approach effectively enhances the feature interactions between relation embeddings and entity embeddings. Simultaneously, we incorporate a priori knowledge-optimized attention mechanism that assigns different contribution weight coefficients to the multiple relation convolution kernels in dynamic convolution, further boosting the expressive power of the model. Extensive experiments on various datasets show that our proposed model consistently outperforms the state-of-the-art baseline methods, with average improvements ranging from 3.28\% to 14.69\% across all model evaluation metrics, while the number of parameters is reduced by 50.66\% to 85.40\% compared to other state-of-the-art models.
\end{abstract}


\section{Introduction}

\IEEEPARstart{K}{nowledge} graphs (KGs) describe real-world facts in the structured form, represented by the triple $(h, r, t)$ containing head entities, relations, and tail entities. KGs provide an indispensable research foundation for knowledge storage, management, and analysis for the development of artificial intelligence. However, KGs generally suffer from incompleteness due to problems such as errors in extraction algorithms or missing data sources. Knowledge embedding is a popular approach to alleviate this challenge, gradually attracting extensive attention from academia and industry. Knowledge graph embedding aims to embed semantic information of research objects into continuous low-dimensional vector spaces, and predict missing links for knowledge graph completion by learning effective representations of entities and relations~\cite{TransO}. This challenging task is also known as link prediction.


Historically, research in knowledge graph embedding has predominantly focused on shallow models relying on translation distance and semantic matching. These models, such as TransE \cite{TransE}, DistMult \cite{DistMult}, and ComplEx \cite{ComplEx}, have laid the foundational work by simplifying the relations between entities and relations through vector space translations and bilinear forms. However, these models exhibit limitations in extracting comprehensive semantic information from knowledge graphs. Specifically, they often struggle to capture the complex and nuanced relations that exist within knowledge graphs, leading to suboptimal performance in tasks such as link prediction and knowledge graph completion.

Recent advancements in natural language processing (NLP), especially the rise of large models based on the Transformer architecture, have attracted significant attention. Some studies have attempted to directly apply these Transformer-based models or other large language models (LLMs) to enhance knowledge graph completion. While these models demonstrate strong capabilities in capturing deep and complex semantic features of triples, their Transformer-based architectures often struggle to effectively model the structural characteristics inherent in KGs. As a result, they may perform relatively well on relation-sparse datasets like WN18RR, where structural demands are limited, but tend to fall short on large-scale and structurally complex KGs \cite{LLM3}. 
Furthermore, their outputs may not fully align with the structured knowledge in the current knowledge graphs \cite{MSHE}. 

Despite significant progress achieved by large models in natural language processing, their application to knowledge graph completion remains challenging. These models, based on a search paradigm, excel at extracting answers from vast and diverse datasets across various domains. However, when dealing with structured knowledge graphs, their effectiveness may be limited compared to specialized knowledge graph embedding models. This is due to the inherent limitations in their architecture, which can lead to hallucinations—predictions that are unsupported by the data \cite{IKGC}. 
As the scale of data increases, the representational capabilities of large models in various fields continue to improve. Nonetheless, the high knowledge coverage inevitably leads to high training costs and issues of inaccuracy and unreliability, without significantly enhancing the quality of low-dimensional triple embeddings or their actual representational capacity \cite{Simre}. Training triple embeddings through neural network methods presents a feasible approach to improving the quality of low-dimensional embeddings.

With the powerful ability of neural networks in different tasks, the deep knowledge embedding model based on convolutional neural networks has gradually attracted more research interest, which can extract more wealthy semantic information and significantly improve the model performance. The advantage of deep knowledge graph embedding models based on convolutional neural networks mainly lies in the fact that convolution can capture the interaction information between entity embeddings and relation embeddings, thus extracting deeper semantic information and substantially improving the expressive power of the model~\cite{InteractE}. Therefore, the key point of such methods mainly centers around the design of the convolution kernel to enhance the degree of interaction between entity embeddings and relation embeddings. 
In particular, in situations where large-scale datasets are not accessible for training, these approaches utilize the local perception and parameter-sharing characteristics inherent in convolutional neural networks.  

ConvE~\cite{ConvE} is a classical convolutional embedding method mainly using a stack reshaping function and external convolution kernel to extract semantic information. Based on ConvE, InteractE~\cite{InteractE} first proposes the chequer reshaping function to improve feature interaction information. ConvR~\cite{ConvR} extracts entity embeddings information with relational embeddings directly acting as an internal convolution kernel to increase feature interaction and improve performance. Moreover, ConvR can achieve more efficient feature extraction with less parameter cost than InteractE, which is more suitable for large-scale datasets. In most existing convolutional embedding methods, to increase the model expressiveness, construct multiple convolution kernels to extract features during the convolution process, and each convolution kernel contributes equally to the feature extraction. However, equal weighting of each convolution kernel is not reasonable in knowledge graph embedding, and each convolution kernel should focus differently on feature extraction. In the knowledge embedding process, although the simple and coarse feature aggregation with the same weight for each convolution kernel can increase the interaction information of the entity embeddings and relation embeddings, it also blurs the contribution of important interaction information. It even introduces some useless noise information to interfere with the embedding performance. 

\begin{figure}[!t]
\centering
\includegraphics[width=\linewidth]{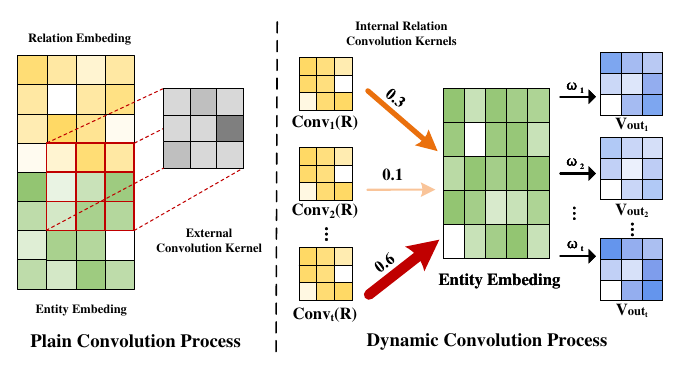}
\caption{Schematic diagram of dynamic convolution and traditional plain convolution process. Dynamic convolution using multiple internal relation convolution kernels differs from the external convolution kernels in plain convolution.}
\label{fig:motivation}
\end{figure}

In this paper, we analyze the advantages and challenges of existing state-of-the-art convolutional embedding methods and propose a novel dynamic convolutional knowledge graph embedding model ConvD, as shown in Figure~\ref{fig:motivation}. ConvD first reshapes the relation vectors into multiple internal convolution kernels to directly extract the entity embedding feature information and improve the degree of model feature interaction. At the same time, the attention mechanism is introduced to calculate the correlation between entity embeddings and relation embeddings, and the adaptive dynamic convolution kernel weight vector is constructed. It should be noted that the dimension of the weight vector of ConvD will be consistent with the number of internal relational convolution kernels and adaptively and dynamically adjusted during the training process. Through this attention enhanced dynamic convolutional interaction, ConvD enables each relational convolution kernel to be effectively and rationally utilized, highlighting the critical feature information and alleviating the interference of useless information. Entity embeddings and relation embeddings can also be realized to interact effectively and adequately, significantly improving the expressive power of the model. Furthermore, a priori knowledge optimization strategy is designed for the traditional attention mechanism to further improve the accuracy of the weight coefficients in the dynamic convolution process. Notably, we construct a 2D priority matrix based on the priori probability of entity-relation pairs in a KG. The higher the priori probability of an entity-relation pair, the higher the attention weight should be assigned to the pair in the dynamic convolution to improve the rationality of capturing the feature interaction information in the knowledge embedding process. In addition, compared to existing convolutional embedding methods, ConvD demonstrates more robust performance, validating the effectiveness of the dynamic convolution strategy. Compared to current representation learning models based on Transformers, ConvD achieves superior performance, or at least comparable performance, with more compact parameters. This will be elaborated upon in detail in the experimental section.

Our contributions are summarized as follows:
\begin{itemize}
    \item We propose a novel dynamic convolutional embedding model ConvD for knowledge graph completion, which directly reshapes the relation embeddings into multiple internal convolution kernels and can effectively enhance feature interactions to improve model performance.
    \item An attention mechanism optimized by priori knowledge is designed, allowing different contribution weight coefficients to be assigned to multiple relation convolution kernels for dynamic convolution, thereby further improving the rationality in the feature extraction process.
    \item Extensive experiments on various datasets demonstrate that the ConvD model consistently outperforms state-of-the-art baseline methods, achieving an average improvement of 8.99\% across all model evaluation metrics. Additionally, the number of parameters is reduced by 68.03\% compared with the best baseline. Ablation experiments validate the effectiveness of the component modules of our model.
\end{itemize}

\section{Related Work}
Existing knowledge graph embedding methods can be classified into three categories: shallow semantic models, transformer-based models, and neural network models.

\subsection{Shallow Semantic Models}

Traditional knowledge graph embedding models assume an inherent correlation: if there exists a certain relation between two entities, their representations in the vector space should be proximate through a specific operation. These models typically interpret relations as transformations in different high-dimensional spaces, such as translations in Euclidean space \cite{TransE}, bilinear transformations \cite{TransR}, rotations in complex space \cite{RotatE}, and reflections in hyperspherical space \cite{TransH}, to extract shallow semantic information. Among these, the TransE model \cite{TransE} was the first to introduce the continuity assumption in knowledge graph embedding, proposing viewing the relation vector as a translation operation from the head entity vector to the tail entity vector. Subsequent works have employed various dimensional spaces, such as hyperplanes \cite{TransH}, Riemannian spheres \cite{5E}, relational mapping spaces \cite{TransD}, multiple mapping spaces \cite{TransG}, and rotational spaces, or have utilized operations like inversion, reflection \cite{TransM}, rotation \cite{RotatE}, and homogeneity to enhance the understanding of complex relational patterns by the model, thus making the representation of relations less rigid.

Another class of classical embedding methods employs tensor decomposition to compute the confidence levels of triples \cite{RESCAL}. Using low-rank tensor decomposition, the relation matrix is decomposed into two low-rank matrices \cite{DistMult}, representing the semantic connections between two entities in a triple \cite{HolE}. Subsequent works have further explored simplifying the complex patterns of relation matrices \cite{ComplEx} and addressing issues such as asymmetric relation matching \cite{SimplE}. The latest work, TuckER \cite{TuckER}, employs a three-dimensional tensor decomposition method to learn vector representations of entities and relations, achieving promising results.

These models can effectively capture shallow semantic information within triples, typically using relation embeddings as a bridge for entity transformation. They are characterized by their simplicity, intuitiveness, and computational efficiency \cite{IncDE}. However, when dealing with knowledge graphs that contain multiple or nonlinear relations, these models often struggle to accurately capture deep semantic information, leading to suboptimal prediction performance~\cite{HJE}.

\subsection{Transformer-based Models}
With the breakthrough of the transformer architecture in the field of natural language processing, many research efforts in the field of knowledge graphs have introduced the transformer architecture to downstream tasks such as link prediction. Some researchers have even directly fine-tuned large models and then applied them to representation learning. As an example, KGT5~\cite{KGT5}, which uses the transformer generic structure of encoder-decoder as a scalable knowledge graph embedding model, transforms the knowledge graph link prediction problem into a sequence-to-sequence task, with remarkable results on large-scale datasets. In addition, KGT5 has carried out a series of useful explorations in the field of knowledge quizzing based on knowledge graph. The Hierarchical transformer model HittER~\cite{HittER}, which integrates source entity neighbourhood features and entity role embedding features, is able to fully take into account the relation context and the information of the source entities themselves according to the balance of different datasets. This approach provides a hierarchical and more nuanced way of modelling entities and relations in the Knowledge Graph.

LLMs have achieved significant success in text understanding and generation \cite{LLM1}, but autoregressive pre-training has made them prone to hallucinations \cite{LLM3}.
Some studies have employed large models to encode textual descriptions of entities and relations to enrich the representation of the Knowledge Graph. Pretrain-KGE~\cite{PretrainKGE} is a representative work, which first obtains textual representations of entities $h, t$ and relations $r$, and then generates representations of the entities and relations from the corresponding text by exploiting the powerful capabilities of LLMs. 
TEA-GLM employs LLMs as zero-shot learners for graph machine learning tasks \cite{TEA-GLM}, while KGEditor converts paths into Chain-of-Thought prompts for multi-hop link prediction \cite{KGEditor}. KoPA, the current state-of-the-art, introduces a knowledge prefix adapter to align structural embeddings with LLM tokens, effectively integrating KG information into LLMs \cite{KoPA}.
Compared to the transformer architecture, convolutional neural networks have efficient computational performance and are more effective when the data size is not particularly large. In addition, the inductive bias in convolutional neural networks is more appropriate for knowledge graph data, while the architecture of multi-layer self-attention mechanism relies more on massive data to learn features, so convolutional neural networks may be more suitable for knowledge graph related tasks, especially when the data size is relatively small to achieve better results.

\subsection{Neural Network Models}

In recent years, researchers have explored the application of convolutional neural networks (CNNs) in knowledge graph embedding models, leveraging the characteristics of CNNs \cite{SACN}, such as local perception and parameter-sharing, to capture complex patterns between entities and relations and predict missing information \cite{ATTH}. These models typically convert entity and relation vectors into 2D matrices and perform feature aggregation by concatenating the head entity and relation embedding matrices. The cross-entropy loss function is then used to determine the correct class of the tail entity \cite{ConvE}. Optimization improvements focus on enhancing the interaction between entities and relations. ConvKB \cite{ConvKB} delves into the global relations between entity entries and relations with the same embedding dimensions to enhance their interaction. InteractE \cite{InteractE} strengthens interactive information by altering feature combinations and using circular convolution during feature extraction. AcrE \cite{AcrE} promotes the interaction of entities and relations at the convolutional level by combining two different types of convolutions. ConvR \cite{ConvR} models the interactions between different positions of entities and relations for the first time. There are also improvements in other areas that promote link prediction performance. R-GCN \cite{RGCN}, a graph-based deep neural network model, utilizes the adjacency information of each entity. SACN \cite{SACN} combines GCN and ConvE to enhance its expressive power. HyperMLN \cite{HyperMLN} is an advanced interpretable knowledge embedding framework that integrates Markov Logic Networks and the Variational EM optimization algorithm for knowledge inference. HyConvE \cite{HyConvE} combines 2D and three-dimensional convolutions for joint training, capturing deep interactions and intrinsic pattern information between entities and relations, thereby improving model representation. 

These models generally enhance feature extraction capabilities through efficient convolutional structures and well-designed combinations of entity and relation embedding features, offering significant advantages in semantic feature learning \cite{TracKGE}. However, the plain convolution process, which involves the use of simple and coarse feature aggregation with the same weights, can obscure the contributions of important interactive information and introduce irrelevant noise, thereby impairing the expressive power of the model \cite{KGEditor}. 

\begin{figure*}[!t]
\centering
\includegraphics[width=166mm]{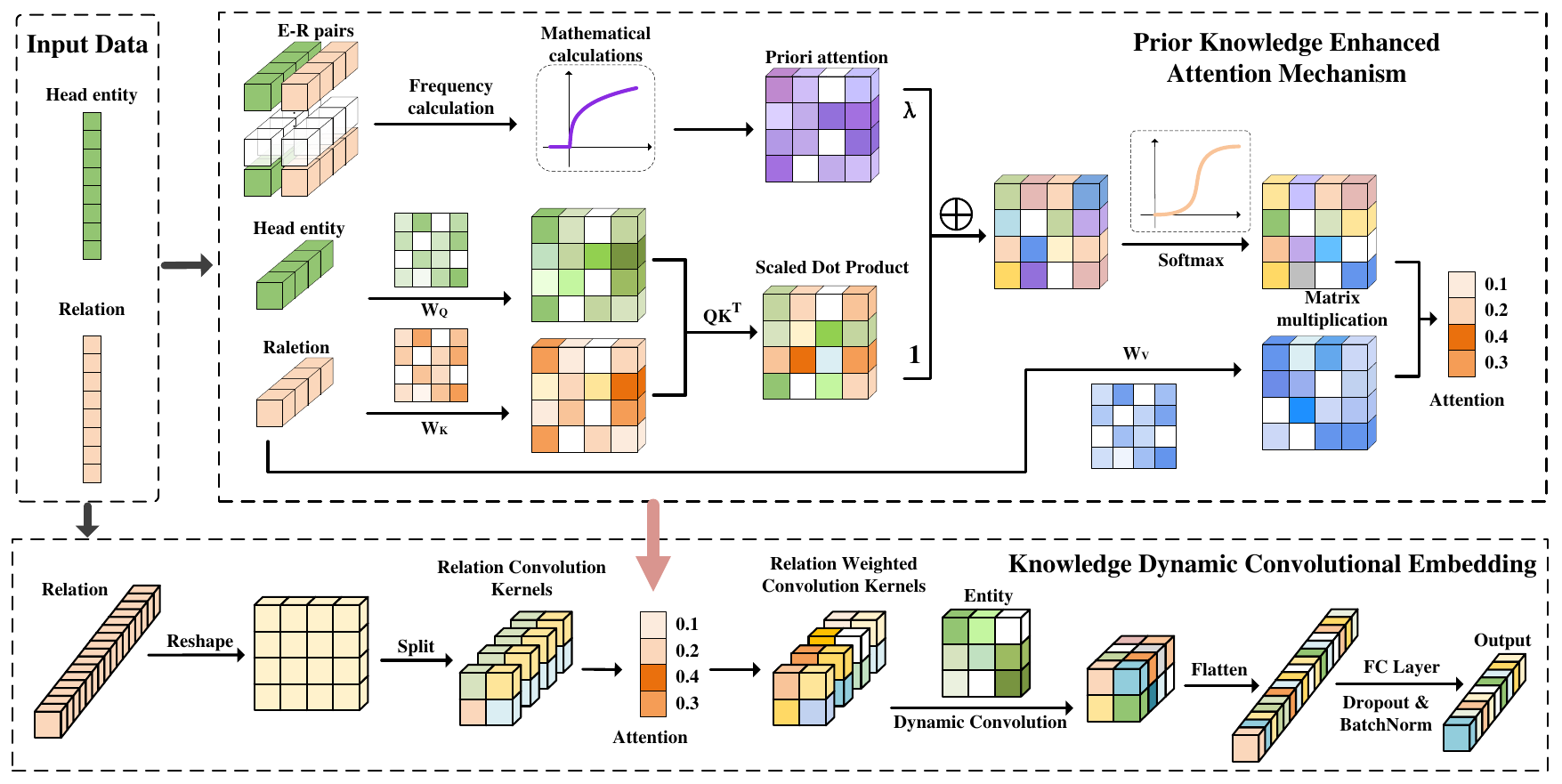}
\caption{The overall framework of the ConvD model.}
\label{fig:model}
\end{figure*}

\section{Preliminaries}
\subsection{Knowledge Graph}
Let \( \mathcal{G} = (\mathcal{E}, \mathcal{R}, \mathcal{T}) \) represent a knowledge graph, where \( \mathcal{E} \) denotes the set of entities, \( \mathcal{R} \) represents the set of relations, and \( \mathcal{T} \) is the set of triples. Specifically, \( \mathcal{T} \) is defined as \( \mathcal{T} = \{(h, r, t) \mid h, t \in \mathcal{E}, r \in \mathcal{R} \} \subseteq \mathcal{E} \times \mathcal{R} \times \mathcal{E} \), where each triple \( (h, r, t) \) consists of a head entity \( h \), a relation \( r \), and a tail entity \( t \). We use \(e_h \) to denote the vector corresponding to the head entity \( h \), and similarly, \(e_r \) and \(e_t \) to represent the vectors corresponding to the relation \( r \) and the tail entity \( t \), respectively.

\subsection{Knowledge Graph Completion} 

Knowledge graph completion aims to predict missing entities or relations within a knowledge graph. Specifically, the entity prediction completion task is given the head (tail) entity and relation to predict the missing tail entity $(h, r, ?)$ or head entity $(?, r, t)$. The relation prediction completion task is given a head and tail entity to predict the missing relation between them, i.e., $(h, ?, t)$. The performance of knowledge graph completion models is typically evaluated through link prediction tasks. In real-world knowledge graph datasets, the number of entities usually far exceeds the number of relations. Therefore, the entity prediction task is more suitable for accurately assessing the effectiveness of model completion, and it is also more commonly utilized.

\subsection{Plain Convolution Process}

The conventional convolution process involves directly concatenating entity embeddings $E$ with relation embeddings $R$, as illustrated in Fig~\ref{fig:motivation}. This concatenation can be mathematically represented as the formation of a feature matrix $M_{ij} = \text{concat}(e_i, r_j)$, where $e_i \in \mathbb{R}^{d_e}$ and $r_j \in \mathbb{R}^{d_r}$ denote the embeddings of entities and relations, respectively, with $d_e$ and $d_r$ representing their respective dimensionalities. Subsequently, independent convolutional kernels are applied to the resulting matrix to extract interaction features within the triplets. Specifically, the convolution operation $f_{conv}$, parameterized by weights $W$ and biases $b$, is employed as follows:

\begin{equation}
    h_{ij} = f_{conv}(M_{ij}; W, b)
\end{equation}

This process effectively captures interactive features between entities and relations, enabling the extraction of deep semantic associations within the knowledge graph. However, this method imposes a dimensionality constraint, requiring $d_e = d_r$ to ensure the feasibility of the convolution operation. This constraint potentially hampers the full utilization of relation-specific features, thereby limiting the model's expressiveness. Furthermore, the interaction between entity and relation embeddings can be further enhanced by introducing an interaction function $f_{enhance}(e_i, r_j)$ that augments the initial concatenation:

\begin{equation}
    g_{ij} = f_{enhance}(e_i, r_j)
\end{equation}

The enhanced interaction process has the potential to improve the model's ability to capture internal interactions, thereby enabling it to effectively discern the complex structural information inherent in the knowledge graph.

\subsection{Priori Knowledge}

By conducting frequency and statistical analyses of knowledge graphs, it is possible to uncover co-occurrence patterns of entities and relations, thereby providing additional a priori knowledge for constraining models. Let $E = {e_1, e_2, ... , e_n}$ denote the set of entities and $R = {r_1, r_2, ... , r_n}$ denote the set of relations in a knowledge graph $\mathcal{G}=\left(\mathcal{E}, \mathcal{R}, \mathcal{T} \right)$. The frequency of an entity-relation pair $(e_i, r_j)$ within the graph can be denoted as $f(e_i, r_j)$ , which serves as a priori information. This frequency helps define the basic concepts and characteristics of entities and relations, thereby allowing us to model the probability distribution $P(e_i, r_j)$, which identifies common patterns and regularities within the knowledge graph. Maximizing the likelihood function $\mathcal{L}(\theta) = \max_{\theta} \prod_{(e_i, r_j) \in G} P_{\theta}(e_i, r_j)^{f(e_i, r_j)}$ is crucial for model construction and optimization. This process enhances the accuracy and plausibility of model predictions by incorporating statistical regularities derived from the knowledge graph.


\section{Methodology}
The overall framework of the ConvD model is illustrated in Fig.~\ref{fig:model}. The final output of the model comprises two main components. Firstly, ConvD reshapes the relation vectors into multiple dynamic convolution kernels of the same dimension, enabling the relation embeddings to directly extract feature information from the entity embeddings, thereby significantly enhancing feature interaction capability. During this process, each convolution kernel contributes different weights to the final output, which are dynamically adjusted adaptively during training. This mechanism ensures that the model can more flexibly capture entity features under different relations, thereby improving the accuracy and expressive power of the embedding representations. Secondly, ConvD incorporates an attention mechanism to compute the feature importance of each dynamic convolution kernel, and the dimensionality of the attention weight coefficient vector is adjusted to align with the number of dynamic convolution kernels. Furthermore, ConvD employs a priori knowledge optimization strategy within the attention mechanism to enhance the accuracy of feature interaction information capture. This strategy constructs a 2D priori matrix based on the priori probabilities of entity-relation pairs in the knowledge graph. The higher the priori probability of an entity-relation pair, the greater the attention weight assigned. In this way, ConvD can more judiciously allocate attention weights, thereby improving the accuracy of feature interaction information captured during the knowledge embedding process.

\subsection{Priori Knowledge Enhanced Attention Mechanism}
In the process of knowledge dynamic convolution interaction, the contribution weight coefficients $\alpha_{i}$ of multiple convolution kernels for dynamic convolution are crucial. ConvD introduces an attention mechanism to compute the correlation between entity embeddings and relation embeddings to construct the weight vectors of adaptive dynamic convolution kernels, as follows:
\begin{equation}
    \boldsymbol{Q}=\boldsymbol{W}_{Q}\boldsymbol{e}_{h}
\end{equation}
where $\boldsymbol{W}_{Q} \in \mathbb{R}^{d{e} \times k}$ is a 2D mapping matrix that performs feature mapping on the original entity embeddings to obtain the parameter matrix $Q$.

\begin{equation}
    \boldsymbol{K}=\boldsymbol{W}_{K}\boldsymbol{e}_{r}
\end{equation}
\begin{equation}
    \boldsymbol{V}=\boldsymbol{W}_{V}\boldsymbol{e}_{r}
\end{equation}
where $\boldsymbol{W}_{K} \in \mathbb{R}^{d{r} \times k}$ and $\boldsymbol{W}_{V} \in \mathbb{R}^{d{r} \times m}$ are 2D mapping matrices that extract relevance information and actual information from the original relation embeddings, respectively. These matrices aid the model in establishing the association between entity features and local relational features to determine which information should be attended to.

The attention mechanism essentially constructs a weight distribution, and this attention allocation can also come from other sources. Therefore, ConvD designs a priori knowledge optimization strategy for the traditional attention mechanism. We construct the 2D priority matrix based on the priori probability of entity-relation pairs in the knowledge graph. The higher the priori probability corresponding to an entity-relation pair, the stronger the correlation between them. The priority matrix is defined as:
\begin{equation}
    \boldsymbol{P}(e_{i},r_{j})=\log_{a} \left ( \mathrm{Freq}(e_{i},r_{j})+1 \right ) 
\end{equation}
where the dimension of $\boldsymbol{P}(e_{i},r_{j})$ is $d_{e} \times d_{r}$, $\mathrm{Freq}(\cdot)$ represents the frequency calculation function, and $a>1$. In particular, the priori knowledge calculation process uses logarithmic smoothing to suppress the effect of noisy data while preventing uncomputable and negative probability anomalies.


Combining ontology cardinality information with the attention mechanism guides the model to more effectively utilize the knowledge graph information during the associated feature extraction process. The integration method is as follows:

\begin{equation}
    \mathrm{Attention}(Q,K,V) = \mathrm{softmax} \left ( \frac {\boldsymbol{Q}\boldsymbol{K}^{T}}{\sqrt {d_{k}}} + \lambda \boldsymbol{P} \right ) \boldsymbol{V}
\end{equation}
in this equation, $\mathrm{Attention(\cdot)}$ denotes the attention score calculation function, $d_k$ represents the dimensionality of the embedding vectors, and $\lambda$ is a hyperparameter that adjusts the weight of integrating priori knowledge. We employ the $\mathrm{softmax}(\cdot)$ function to compute the interaction feature correlation probabilities infused with priori knowledge and subsequently calculate the contribution weight coefficients of the multi-dynamic convolution kernels with the $V$ matrix.


\begin{equation}
    \alpha_{i} = \mathrm{Attention}(Q,K,V)
\end{equation}

For simplicity, the contribution weight coefficient of the multi-dynamic convolution kernels is denoted by the contribution weight factor $\alpha_{i}$, which corresponds to the result of the aforementioned calculation $\mathrm{Attention}(Q, K, V)$.

\subsection{Knowledge Dynamic Convolutional Embedding}
Given a knowledge triple $(h, r, t)$, we first construct head entity embedding vectors $\boldsymbol{e}_{h} \in \mathbb{R}^{d_{e}}$ and relation embedding vectors $\boldsymbol{e}_{r} \in \mathbb{R}^{d_{r}}$. Subsequently, ConvD utilizes dynamic convolution with multiple convolution kernels to capture feature interaction information between relation embeddings and entity embeddings. The convolution kernel of ConvD is defined as: 

\begin{equation}
    \boldsymbol{\omega}_{i} = \alpha_{i} \cdot \boldsymbol{\omega}_{r}^{i}
\end{equation}
where $\alpha_{i}$ is the contribution weight factor for each convolution kernel and $\boldsymbol{\omega}_{r}^{i}$ represents each specific convolution kernel. This weight factor measures the degree of correlation between the local relational features and the entity embedding vectors. A higher contribution weight factor indicates a deeper correlation and greater importance in feature interaction. Notably, each convolution kernel of ConvD is constructed directly from the 2D reshaping of the relation embeddings, as follows:
\begin{equation}
    \boldsymbol{\omega}_{r}^{m} = [conv_{1}(\overline{\boldsymbol{e}_{r}}); conv_{2}(\overline{\boldsymbol{e}_{r}}); ...; conv_{m}(\overline{\boldsymbol{e}_{r}})]
\end{equation}
where $conv(\cdot)$ is the convolutional construction function and $m$ is the number of the convolution kernel. Different convolution construction functions can build various relational convolution kernels based on the 2D relational embedding features, thereby capturing distinct local relational features. The dimension of each convolution kernel $\boldsymbol{\omega}_{r}^{i}$ is set to $r_{w} \times r_{h}$, and $\overline{\boldsymbol{e}_{r}}$ is the 2D reshaping of the relation embeddings with dimension $r_{w}\sqrt{m} \times r_{h}\sqrt{m}$. Expressly, $m$ must be the square of an integer greater than zero to ensure that $\sqrt{m}$ is an integer.

The core of enhancing feature capture and expression capabilities lies in thoroughly exploring the correlation information inherent in dynamic convolution kernels and effectively integrating it into the entity features. ConvD uses multiple relation convolution kernels dynamically convolved with the 2D reshaping of the head entity embedding vectors. Then, the feature vector is output through a fully connected layer.
\begin{equation}
    \boldsymbol{v}_{out} = \mathrm{FC} \left (\sum_{i=1}^{m} \overline{\boldsymbol{e}_{h}} * \boldsymbol{\omega}_{i}  \right ) 
\end{equation}
where $\boldsymbol{v}_{out}$ is the output feature vector, $*$ denotes the convolution operator, and $\overline{\boldsymbol{e}_{h}} \in \mathbb{R}^{d_{w} \times d_{h}}$ represents the 2D reshaping of the head entity embedding vectors. Likewise, the dimension of the entity embedding vectors $\boldsymbol{e}_{h}$ is $d_{e}=d_{w}d_{h} \times 1$ and the dimension of the relation embedding vectors $\boldsymbol{e}_{r}$ is $d_{r}=mr_{w}r_{h} \times 1$.

\subsection{Joint Training}

\begin{figure}[H]
    \begin{algorithm}[H]
        \caption{Training procedure for ConvD}
        \label{algorithm1}
        \textbf{Input}: Observed tuples $\mathcal{G}=\left(\mathcal{E}, \mathcal{R}, \mathcal{T} \right)$, iteration count $n_{\mathrm{iter}}$, mini-batch size $m_{b}=128$\\
        \textbf{Output}: The score of each tuple
        \begin{algorithmic}[1]
                \STATE Sample a mini-batch $\mathcal{G}_{\mathrm{batch}} \in \mathcal{G}$ of size $m_{b}$.
                
                \FOR{each tuple $g \in \mathcal{G}_{\mathrm{batch}}$}
                    \STATE Constructing 2D embedding matrix of $g$;
                    // Reshape the 1D vector into a 2D matrix.
                    \STATE $\mathbf{P}\leftarrow$ compute priori knowledge using (4);
                    \STATE $\boldsymbol{\alpha_ {i}}\leftarrow$ compute attention weight using (5);
                    // The attention coefficients are aligned using priori knowledge.
                    \STATE $\mathbf{v}_{out} \leftarrow$ compute dynamic convolution results using (9);
                    \STATE $\mathrm{score}\leftarrow$ get the final score of the tuple using (10);
                \ENDFOR
                
                \STATE Update learnable parameters w.r.t. gradients based on the whole objective in (11).
            
            \STATE \textbf{return} solution
        \end{algorithmic}
    \end{algorithm}
\end{figure}

Algorithm~\ref{algorithm1} summarizes the training procedure for ConvD. When training model parameters, unlike the 1-1 method that treats each triple as an independent training sample, ConvD employs the 1-N training method to simultaneously train with multiple candidate entities, thereby accelerating the computation process. During each training iteration, the model can compute the scores for all candidate entities at once, rather than calculating them individually. This significantly speeds up the training process and allows the model to learn global relational patterns more quickly, reducing the overall training time required. In this study, $(h, r, ?)$ is computed, with the results represented by a vector whose dimension corresponds to the number of entities in the dataset.
\begin{equation}
    \psi(h,r,t) = f\left( \mathrm{ReLU}\left( \boldsymbol{v}_{out} \right) \boldsymbol{W}+\boldsymbol{b} \right) \boldsymbol{e}_{t}
\end{equation}
where $\boldsymbol{v}_{out}$ is the output result of the dynamic convolution operation and $\boldsymbol{e}_{t}$ represents the object entity embedding matrix. $\boldsymbol{W}$ is the feature mapping matrix, $\boldsymbol{b}$ is the bias term, and $f(\cdot)$ denotes the \textit{sigmoid} function.

Moreover, ConvD uses three dropout layers to prevent overfitting for better training results, specifically dropout in the input data part, vector reshaping representation part, and fully connected layer part, respectively. Besides, ConvD utilizes \textit{batch normalization} to speed up the dynamic convolution operation and uses the \textit{Adam optimizer} and \textit{label smoothing} method to train the model. The knowledge graph link prediction task can be conceptualized as a classification problem, wherein the objective is to determine whether a specific relation exists between two entities. The number of classes in this classification task corresponds to the number of distinct entity types within the entity set. Consequently, the model is trained using the cross-entropy loss function.  
\begin{equation}
\mathcal{L}(h,r)=-\frac {1}{|\varepsilon|}\sum_{o\cup \varepsilon}y_t\log(\rho)+(1-y_t)\log(1-\rho)
\end{equation}
where $\varepsilon$ is the number of candidate entities and $\rho$ denotes the model score $\psi(h,r,t)$. The $y_t$ is a label, $y_t=1$ if $(h,r,t)$ is the correct triple, otherwise $y_t=0$.

To ensure the model attains the optimal hyperparameter configuration during training, a hyperparameter tuning strategy combining grid search and Bayesian optimization is employed. Grid search systematically explores all possible hyperparameter combinations and selects the best configuration through evaluation. Given the complexity of the hyperparameter space and computational costs, Bayesian optimization is further introduced. 
\begin{equation}
\mathcal{H}_{\text{best}} = \arg\max_{\mathcal{H} \in \mathcal{H}_{\text{grid}}} \left( \max_{\mathcal{H}^{\prime} \in \mathcal{H}_{\text{Bayes}}} \mathbb{E}(\mathcal{H}^{\prime}) \right)
\end{equation}
where the symbol \(\mathcal{H}_{\text{best}}\) represents the optimal hyperparameter configuration identified through the optimization process. The set \(\mathcal{H}_{\text{grid}}\) denotes the initial pool of hyperparameter candidates derived from a grid search procedure. Subsequently, \(\mathcal{H}_{\text{Bayes}}\) refers to the refined set of hyperparameter candidates that have undergone further adjustment via Bayesian optimization techniques. The function \(\mathbb{E}( \cdot )\), which stands for Expected Improvement, serves as a metric to evaluate the efficacy of the Bayesian optimization, where \(\mathcal{H}'\) signifies an individual hyperparameter configuration within the Bayesian optimization process.

This method constructs a probabilistic model between the hyperparameters and model performance, intelligently narrowing the search space, accelerating the optimization process, and reducing computational overhead. Additionally, to prevent overfitting, early stopping is implemented during training. When the performance on the validation set no longer improves significantly, training is automatically terminated, thereby avoiding unnecessary computations and mitigating the risk of overfitting. This ensures that the model improves both its performance on the training set and its generalization ability to unseen data.

The time complexity of ConvD is primarily dominated by three key components: the generation of dynamic convolutional kernels, the computation of the attention mechanism, and the convolutional operations. The generation of dynamic convolutional kernels has a time complexity of \(\mathcal{O}(m \cdot d_r)\). The attention mechanism computation involves interactions between entity and relation embeddings, resulting in a complexity of \(\mathcal{O}(d_e \cdot d_r)\). Finally, the convolutional operations contribute a complexity of \(\mathcal{O}(m \cdot k^2 \cdot d_e)\). The overall time complexity can be expressed as \( \mathcal{O}(m \cdot d_r) + \mathcal{O}(d_e \cdot d_r) + \mathcal{O}(m \cdot k^2 \cdot d_e) \).

\section{Experiments}

This section provides a detailed description of the experimental procedures. We first introduce the datasets used in the experiments and the classical baseline methods employed. Following this, we explain the metrics used for evaluation. We then present a thorough account of the experimental configurations and analyze the model's performance based on the results of the link prediction task. Additionally, we conduct ablation studies and parameter comparison experiments to provide an in-depth analysis of dynamic convolution.

\subsection{Datasets and Baselines}
\subsubsection{Datasets}
This paper uses three publicly available datasets for experiments. FB15K-237~\cite{FB15K237} and WN18RR~\cite{WN18RR} are the most common datasets in knowledge graph embedding and can be employed to compare the experimental performance of many related works. It is worth noting that the WN18RR dataset, with only 11 relations, performs poorly in convolution-based models. In addition, we used the DB100K~\cite{DB100K} dataset to verify the robustness of the ConvD model, which is a competitive large-scale dataset. The data size of DB100K is close to the size of the YAGO3-10 dataset. Still, the number of relations is 12.7 times more than that of YAGO3-10, which is a good choice for validating the generalization ability of the relation-centered knowledge graph embedding model. The detailed statistics of the datasets are summarized in Table~\ref{table:datasets}.

\begin{table}[ht]
    \begin{center}
        \caption{Dataset Statistics}
        \label{table:datasets}
          \begin{tabular}{cccccc}
          \toprule  
          \textbf{Dataset} & \textbf{\#Rel}  &  \textbf{\#Ent} & \textbf{\#Train} & \textbf{\#Valid} & \textbf{\#Test} \\ 
          \midrule 
          FB15K-237 & 237  &  14,541 & 272,115 & 17,535 & 20,466 \\
          WN18RR & 11  &  40,943 & 86,835 & 3,034 & 3,134 \\
          DB100K & 470  &  99,604 & 597,572 & 50,000 & 50,000 \\
          \bottomrule 
          \end{tabular}
    \end{center}
    \thanks{The size of the \#Train, \#Valid, and \#Test columns represent the number of tuples, respectively.}
\end{table}

\begin{table*}[ht]
    \begin{center}
    \caption{Link Prediction Results}
    \label{table:FB15B237-WN18RR}
    \resizebox{\textwidth}{!}{
    \begin{tabular}{c|c|cccc|cccc|cccc}  
        \hline
        \multirow{2}{*}{\textbf{Category}} & \multirow{2}{*}{\textbf{Model}} & \multicolumn{4}{c|}{\textbf{FB15K-237}} & \multicolumn{4}{c|}{\textbf{WN18RR}} & \multicolumn{4}{c}{\textbf{DB100K}} \\
        \cline{3-14}
        ~ & ~ & \textbf{MRR} & \textbf{Hits@10} & \textbf{Hits@3} & \textbf{Hits@1} & \textbf{MRR} & \textbf{Hits@10} & \textbf{Hits@3} & \textbf{Hits@1} & \textbf{MRR} & \textbf{Hits@10} & \textbf{Hits@3} & \textbf{Hits@1} \\
        \hline
        \multirow{4}{*}{\makecell{Shallow\\Semantic\\Models}} & TransE~\cite{TransE} & 0.287 & 0.475 & 0.325 & 0.192 & 0.193 & 0.445 & 0.370 & 0.003 & 0.111 & 0.270 & 0.164 & 0.016\\
        ~ & DistMult~\cite{DistMult} & 0.241 & 0.419 & 0.263 & 0.155 & 0.430 & 0.490 & 0.440 & 0.390 & 0.233 & 0.448 & 0.301 & 0.115\\
        ~ & ComplEx~\cite{ComplEx} & 0.247 & 0.428 & 0.275 & 0.158 & 0.440 & 0.510 & 0.460 & 0.410 & 0.242 & 0.440 & 0.312 & 0.126 \\
        ~ & RotatE~\cite{RotatE} & 0.338 & 0.533 & 0.375 & 0.241 & 0.476 & \textbf{0.571} & 0.492 & 0.428 & 0.352 & 0.526 & 0.381 & 0.258 \\

        \hline
             
        \multirow{12}{*}{\makecell{Neural\\Network\\Models}} & Neural LP~\cite{NeuralLP} & 0.240 & 0.362 & - & - & - & - & - & - & - & - & - & - \\
        ~ & R-GCN~\cite{RGCN} & 0.248 & 0.417 & 0.258 & 0.153 & 0.123 & 0.207 & 0.137 & 0.080 & - & - & - & - \\
        ~ & ConvE~\cite{ConvE} & 0.316 & 0.491 & 0.350 & 0.239 & 0.460 & 0.480 & 0.430  & 0.390 & 0.325 & 0.530 & - & 0.236  \\
        ~ & ConvKB~\cite{ConvKB} & \underline{0.396} & 0.517 & - & - & 0.249 & 0.524 & 0.417 & 0.057 & 0.346 & 0.549 & - & 0.249 \\
        ~ & ConvR~\cite{ConvR} & 0.350 & 0.528 & 0.385 & 0.261 & 0.475 & 0.537 & 0.489 & 0.443 & 0.362 & 0.560 & - & 0.265 \\
        ~ & SACN~\cite{SACN} & 0.350 & 0.540 & - & 0.260 & 0.463 & 0.534 & 0.478 & 0.429 & - & - & - & -\\
        ~ & InteractE~\cite{InteractE} & 0.354 & 0.535 & - & 0.263 & 0.463 & 0.528 & - & 0.430 & 0.364 & 0.554 & - & 0.253 \\
        ~ & AcrE~\cite{AcrE} & 0.358 & 0.545 & 0.393 & 0.266 & 0.463 & 0.534 & 0.478 &  0.429 & \underline{0.413} &  \underline{0.588} & \textbf{0.472} & \underline{0.314}\\
        ~ & DT-GCN~\cite{DT-GCN} & 0.321 & 0.430 & 0.278 & 0.167 & 0.322 & 0.471 & 0.340 & 0.193 & 0.287 & 0.427 & 0.311 & 0.187\\
        ~ & HyperMLN~\cite{HyperMLN} & 0.386 & 0.513 & 0.379 & 0.245 & 0.435 & 0.521 & 0.461 & 0.410 & 0.319 & 0.514 & 0.360 & 0.221 \\
        ~ & SAttLE~\cite{SAttLE} & 0.360 & 0.445 & 0.396 & 0.268 & 0.476 & 0.540 & \underline{0.508} & 0.442 & 0.309 & 0.427 & 0.334 & 0.257 \\
        ~ & HyConvE~\cite{HyConvE} & 0.339 & 0.458 & - & 0.212 & 0.461 & 0.534 & - & 0.432 & 0.347 & 0.549 & 0.349 & 0.245\\

        \hline

        \multirow{3}{*}{\makecell{Transformer\\-base\\Models}} & Pretrain-KGE~\cite{PretrainKGE} & 0.350 & 0.554 & - & - & 0.459 & 0.553 & - & - & - & - & - & -\\
        ~ & HittER~\cite{HittER} & 0.373 & \underline{0.558} & \underline{0.409} & \underline{0.279} & \underline{0.503} & \underline{0.584} & \textbf{0.516} & 0.462 & 0.384 & 0.561 & 0.353 & 0.285 \\
        ~ & KGT5~\cite{KGT5} & 0.276 & 0.414 & 0.376 & 0.210 & \textbf{0.508} & 0.544 & 0.501 & \textbf{0.487} & 0.326 & 0.485 & 0.352 & 0.274\\
        
        \hline
        ~ & \textbf{ConvD (Ours)} & \textbf{0.409} & \textbf{0.588} & \textbf{0.447}  & \textbf{0.320} & 0.479 & \textbf{0.587} & 0.495 & \underline{0.465} & \textbf{0.415} & \textbf{0.607} & \underline{0.467} & \textbf{0.315}\\
        \hline
    \end{tabular}
    }
    \end{center}
    \thanks{Link prediction results for FB15K-237, WN18RR, and DB100K. The best results are in boldface and the second best are underlined. For DT-GCN, HyperMLN, and Pretrain-KGE models, since the original paper did not provide the experimental results of FB15K-237 and WN18RR datasets, we use the source code to supplement in the local environment. The results for the large-scale dataset DB100K are taken from~\cite{AcrE}. For RotatE, ConvR, ConvKB, InteractE, DT-GCN, HittER, HyperMLN, KGT5, SAttLE, and HyConvE models, since the original paper did not provide the experimental results of DB100K dataset, we use the source code to supplement in the local environment.}
\end{table*}

\subsubsection{Baselines}
We compared the proposed model ConvD with \textbf{19} classical and state-of-the-art baseline methods, which can be roughly divided into three categories: shallow semantic models, transformer-based models and neural network models.
\begin{itemize}
    \item \textbf{Shallow Semantic Models:} TransE, DistMult, ComplEx, and RotatE.
    \item \textbf{Transformer-based Models:} pretrain-KGE, HittER, and KGT5.
    \item \textbf{Neural Network Models:}Neural LP, R-GCN, ConvE, ConvKB, ConvR, SACN, AcrE, InteractE, DT-GCN, HyperMLN, HyConvE, and SAttLE.
\end{itemize}

\subsection{Evaluation Metrics}

The process for the link prediction task involves a given triple $(h,r,t)$, where either the head entity $h$ or the tail entity $t$ is hidden, resulting in the form $(h,r,?)$ or $(?,r,t)$. The objective is to predict the hidden entity. In link prediction, the goal is not merely to find the best answer, but to evaluate the ranking of the correct triples among the candidate triples in the test task. The evaluation process is based on the method provided by the TransE model: for each triple $(h,r,t)$ in the test dataset $T$, either the head or tail entity is removed, and every entity $e \in E$ in the knowledge graph is used to replace it. The corresponding scores are then calculated using a scoring function. Candidate triples are ranked in ascending order based on their scores, and the results are evaluated using specific metrics. The link prediction task employs the following two evaluation metrics:

Mean Reciprocal Rank (MRR): This metric is the average of the reciprocal ranks of each correctly predicted triple in the test set, highlighting the importance of correctly predicted triples that are ranked highly. The MRR is computed using the following formula:

\begin{equation}
    \mathrm{MRR} = \frac{1}{\sum_{h \in E_{test}} |r|} \sum_{h \in E_{test}} \sum_{p=1}^{|r|} \frac{1}{\mathrm{rank_{p}}(h)}
\end{equation}

where $E_{test}$ represents the set of test triples, and $\mathrm{rank_{p}}(h)$ denotes the rank of the correct triple among the candidate triples after replacing the entity. MRR, as an evaluation metric, aims to encourage the model to predict correct triples with high confidence.

Hit Ratio ($\mathrm{Hits@N}$): This metric measures whether the correct triple is included among the top N candidate triples. Common values include $\mathrm{Hits@1}$, $\mathrm{Hits@3}$, and $\mathrm{Hits@10}$. The calculation formula is as follows:

\begin{equation}
    \mathrm{Hits@N} = \frac{\sum_{h \in E_{test}} \sum_{p=1}^{|r|} \mathrm{cond}(\mathrm{rank_{p}}(h) \leq k)}{\sum_{h \in E_{test}} |r|}
\end{equation}

\noindent the value of $\mathrm{Hits@1}$, $\mathrm{Hits@3}$, and $\mathrm{Hits@10}$ respectively. $\mathrm{Hits@N}$ provides a clear threshold for evaluating whether the model can successfully identify the correct triple among a given number of candidates, thereby making the task requirements more explicit.

\subsection{Experimental Setup}
\subsubsection{Evaluation Settings}
We use link prediction as the proposed model evaluation task. Specifically, for each triple $(h, r, t)$ of the test set, $t$ in the triple is replaced using each entity in the set of entities to generate corrupted triples, and a scoring function computes the scores of all corrupted triples. Since the constructed corrupted triples may also be correct, this paper adopts the filter setting \cite{ConvR}. The low-dimensional embeddings for each entity and relation are initialized with random values. It is worth stating that the performance of most of the baseline methods in the experimental section is taken from the original paper, and the experimental results with ``-'' are results that were not presented in the original paper. We have only performed partial local implementations of recent models with open-source code to supplement the necessary experimental evaluation metric values.

\begin{table*}[ht]
    \begin{center}
      \caption{Parameter Validity}
      \label{tab:parameter-vality}  
        \begin{tabular}{c|ccc|ccc|ccc}
        \toprule
        \multirow{2}{*}{\textbf{Model}} & \multicolumn{3}{c|}{\textbf{FB15K-237}} & \multicolumn{3}{c|}{\textbf{WN18RR}} & \multicolumn{3}{c}{\textbf{DB100K}} \\
        \cline{2-10}
         & \textbf{MRR} & \textbf{Hits@10} & \textbf{Parameters} & \textbf{MRR} & \textbf{Hits@10} & \textbf{Parameters} & \textbf{MRR} & \textbf{Hits@10} & \textbf{Parameters} \\
        \hline
        HittER~\cite{HittER} & 0.373 & 0.558 & 11.20M & 0.503 & 0.584 & 10.89M & 0.384 & 0.561 & 89.61M \\
        KGT5~\cite{KGT5} & 0.276 & 0.414 & 16.21M & \textbf{0.508} & 0.544 & 14.25M & 0.326 & 0.485 & 126.43M \\   
        ConvE~\cite{ConvE} & 0.316 & 0.491 & 5.05M & 0.460 & 0.480 & 3.84M & 0.325 & 0.530 & 33.86M \\
        InteractE~\cite{InteractE} & 0.354 & 0.535 & 6.12M & 0.463 & 0.528 & 4.29M & 0.364 & 0.554 & 39.67M \\
        \hline
        ConvD (Ours) & \textbf{0.409} & \textbf{0.588} & \textbf{2.65M} & 0.479 & \textbf{0.587} & \textbf{1.95M} & \textbf{0.415} & \textbf{0.607} & \textbf{19.34M} \\
        \bottomrule
      \end{tabular}
    \end{center}
\end{table*}

\begin{table*}[ht]
    \begin{center}
     \caption{Results of ablation study}
     \label{table:ablation-study}
     \setlength{\tabcolsep}{4pt}
        \begin{tabular}{c|cccc|cccc|cccc}  
            \hline
            \multirow{2}{*}{\textbf{Model}} & \multicolumn{4}{c|}{\textbf{FB15K-237}} & \multicolumn{4}{c|}{\textbf{WN18RR}} & \multicolumn{4}{c}{\textbf{DB100K}} \\
            \cline{2-13}
            ~ & \textbf{MRR} & \textbf{Hits@10} & \textbf{Hits@3} & \textbf{Hits@1} & \textbf{MRR} & \textbf{Hits@10} & \textbf{Hits@3} & \textbf{Hits@1} & \textbf{MRR} & \textbf{Hits@10} & \textbf{Hits@3} & \textbf{Hits@1} \\
            \hline
            w/o Priori Knowledge & 0.403 & 0.575 & 0.434 & 0.312 & 0.473 & 0.541 & 0.482 & 0.440 & 0.401 & 0.589 & 0.455 & 0.309 \\
            w/o Attention  & 0.365 & 0.534 & 0.397 & 0.274 & 0.385 & 0.467 & 0.458 & 0.353 & 0.367 & 0.550 & 0.425 & 0.301\\
            w/o Priori\&Attention & 0.358 & 0.529 & 0.387 & 0.267 & 0.359 & 0.449 & 0.438 & 0.325 & 0.360 & 0.547 & 0.421 & 0.301 \\
            \hline
            \textbf{ConvD} & \textbf{0.409} & \textbf{0.588} & \textbf{0.447} & \textbf{0.320} & \textbf{0.479} & \textbf{0.554} & \textbf{0.495} & \textbf{0.447} & \textbf{0.415} & \textbf{0.607} & \textbf{0.467} & \textbf{0.315} \\
            \hline
        \end{tabular}
    \end{center}
\end{table*}

\subsubsection{Parameter Settings}
The ConvD model is implemented on the PyTorch architecture. We set the batch size of each training to 128, the initial learning rate to 0.003, and the label smoothing coefficient to 0.1, and use hyperparameter tuning strategy combining grid search and Bayesian optimization to optimize the hyperparameter combinations. Furthermore, the dimension of the entity embeddings are tuned in the range of $d_{e} \in \{100, 150, 200, 250, 300\}$, and the priori knowledge weighting coefficients are tuned in the range of $\lambda \in \{0.1, 0.2, 0.3, 0.4\}$. We use the average results of ten runs as the final performance, and the following experiments all follow this setup. On each dataset, we select the optimal hyperparameter configuration with the highest MRR on the validation set within 200 epochs and report its performance on the test set.

The detailed implementation information of the ConvD model is available at this GitHub link~\footnote{\url{https://github.com/xiumu-gg/ConvD}}.

\subsection{Link Prediction Results}
The results of link prediction experiments of our proposed model on FB15K-237, WN18RR, and DB100K datasets are shown in Table~\ref{table:FB15B237-WN18RR}. From the experimental results, ConvD improves MRR by 3.28\%, Hits@10 by 5.37\%, and Hits@1 by 14.69\% over the optimal baseline method on the FB15K-237 dataset. The average improvement across all evaluation metrics is 8.99\%. ConvD has not shown an advantage on the WN18RR dataset. As noted in the ConvE and InteractE studies, the WN18RR dataset has a relatively low average relation density and is better suited to shallow models, whereas deeper models struggle to achieve excellent performance. We initially deduced that ConvD is categorised as a deep model which employs relational embeddings as a convolutional kernel to enhance the interaction between embeddings. In the WN18RR dataset, ConvD is limited in fully extracting interaction features due to its low relational density and therefore fails to achieve optimal results on this dataset. 

To validate the generalisation ability of the model, we concurrently link the results of the prediction experiments on the DB100K dataset. The experimental results show that the proposed model significantly outperforms the optimal baseline model, proving ConvD has a strong generalization ability on larger-scale datasets. It is worth noting that DB100K is a larger-scale dataset with more relations, which is an experimental validation with practical applications for knowledge graph embedding focusing on relations.



The ConvD model has been experimentally proven to have superior performance on several different public datasets. On all model evaluation metrics, ConvD outperforms or at least equals the performance of current state-of-the-art models. Since ConvD directly reshapes relation embeddings into internal convolutional kernels, ConvD is able to dynamically and adaptively adjust the contribution weight coefficients of multiple convolutional kernels on large-scale datasets with a large number of relations to ensure better generalisation capability and performance through an attention mechanism optimised by a priori knowledge.


\subsection{Parametric Efficiency Comparison}
We select the four state-of-the-art methods for parametric efficiency comparison experiments. Specifically, two models based on the transformer architecture (HittER and KGT5) and two models based on convolutional neural networks (ConvE and InteractE). In TABLE ~\ref{tab:parameter-vality}, we present the results of three benchmarks that show that ConvD performs well in terms of parameter usage efficiency.

Experimental results indicate that ConvD achieves state-of-the-art or superior performance with more compact parameters. The average number of parameters for ConvD is 85.40\% less than the transformer-based model and 50.66\% less than the convolutional neural network model. This further confirms that the convolutional neural network approach has a significant advantage in terms of the number of parameters compared to the transformer-based approach. The knowledge graph representation learning model based on transformer also exhibits outstanding performance. 

However, such methodologies necessitate the configuration of numerous free hyperparameters, resulting in models characterized by exceedingly high time and memory complexity. In contrast, convolutional neural network-based knowledge graph representation learning approaches facilitate representation learning through the sharing of free parameters, enhancing scalability, particularly at smaller embedding dimensions. These challenges underscore the inapplicability of the transformer architecture to knowledge graph representation learning tasks involving intricate semantics and knowledge coupling. 

Simultaneously, experiments have demonstrated that the ConvD model exhibits significantly lower parameter counts compared to other convolutional neural network models, confirming the enhanced effectiveness of dynamic convolution over traditional convolution. During training, ConvD eschews the use of redundant parameters as external convolution kernels and instead directly employs relation embeddings as internal convolution kernels, further validating the superiority of dynamic convolution.

\begin{figure*}[!t]
    \centering
    \includegraphics[width=176mm]{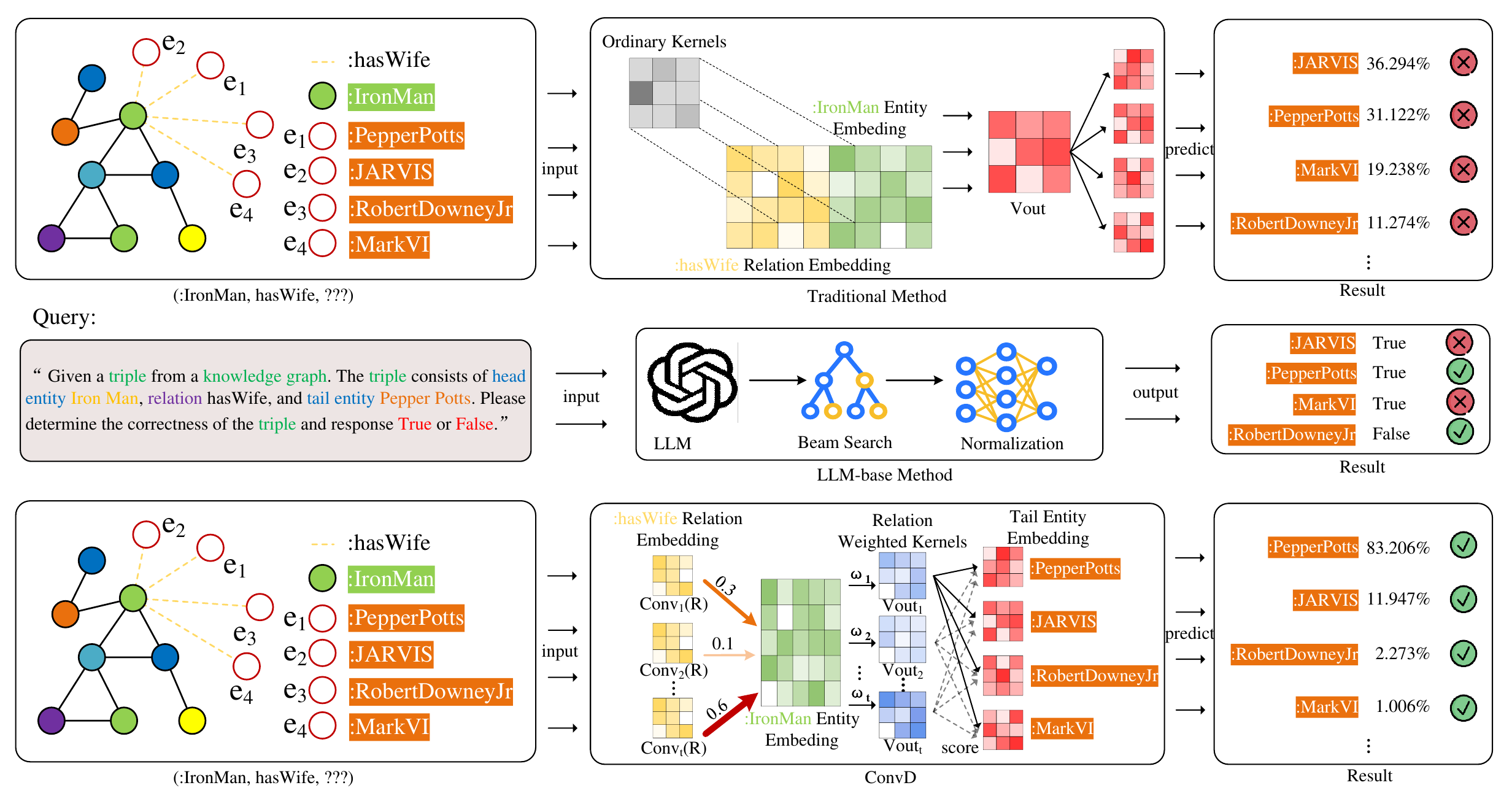}
    \caption{Illustration of Experimental Design for Case Study.}
    \label{fig:case}
\end{figure*}

\subsection{Ablation Study}

Ablation experiment is a necessary research content to verify the function of each module of the knowledge embedding model. We selected three standard knowledge graph datasets, FB15K-237, WN18RR, and DB100K, and conducted ablation experiments using the same hyperparameters as the link prediction experiments, and the results of the experiments are shown in TABLE ~\ref{table:ablation-study}. Especially, we perform three different sets of ablation experiments, namely ConvD without a priori knowledge optimization strategy (w/o Priori Knowledge), ConvD without the attention mechanism (w/o Attention), and ConvD without the priori knowledge-optimized attentional mechanism (w/o Priori\&Attention).

From the experimental results, it can be seen that all the model evaluation metrics of ConvD (w/o Priori Knowledge) have performance reduction ranging from 1.47\% to 2.91\% on the FB15K-237 dataset, from 1.25\% to 2.63\% on the WN18RR dataset, and from 1.90\% to 3.14\% on the DB100K dataset. The integration of priori knowledge enhances the model's predictive accuracy across various datasets, highlighting its non-negligible impact. By guiding the dynamic convolution process with structured domain information, priori knowledge helps the model to better capture meaningful interactions between entities and relations, leading to more precise candidate scoring. ConvD (w/o Attention) all model evaluation metrics degraded performance by 9.18\% to 14.38\% on the FB15K-237 dataset, 7.48\% to 21.03\% on the WN18RR dataset, and 4.45\% to 11.35\% on the DB100K dataset. ConvD (w/o Priori\&Attention) all model evaluation metrics degraded performance by 10.03\% to 16.56\% on the FB15K-237 dataset, 11.52\% to 27.29\% on the WN18RR dataset, and 4.45\% to.13.14\% on the DB100K dataset. Overall, the ablation experiments highlight that each constituent module of ConvD is essential. While the attention mechanism contributes significantly to performance, the role of priori knowledge is also critical, especially in stabilizing and refining the relational representation learning.

\subsection{Case Study Experiment}

To provide a detailed comparison between the dynamic convolution process and the plain convolution process, we extract an example from the FB15K-237 dataset to illustrate the link prediction process using both ConvD and traditional KG embedding models.

We choose Iron Man, a superhero from the Marvel Universe, as the head entity, represented by \texttt{:IronMan}, and predict who his wife is, with the relation denoted as \texttt{:hasWife}. The task involves predicting the tail entity given the known head entity \texttt{:IronMan} and relation \texttt{:hasWife}, which is formulated as the triplet (\texttt{:IronMan}, \texttt{:hasWife}, ?), as shown in Figure ~\ref{fig:case}.

Traditional KG embedding methods transform the entities \texttt{:IronMan} and the relation \texttt{:hasWife} into embedding matrices of the same dimension. These matrices are then concatenated to form the feature context matrices, which ard convolved using ordinary kernels to extract the features of the entities and relations. When the convolution operation reaches the position corresponding to the concatenation of \texttt{:IronMan} and \texttt{:hasWife}, the extracted features represent the interaction between the entity and the relation. Subsequently, these features are integrated and compared with the candidate tail entity embedding matrices to score and calculate the probability of each candidate entity. The experiment results show that the traditional method incorrectly predicts \texttt{JARVIS}, an AI, as Iron Man’s wife, failing to identify \texttt{Pepper Potts} as the correct result.

In contrast, ConvD adopts a different approach by reshaping \texttt{:hasWife} into multiple small-sized embedding matrices and uses these matrices as convolutional kernels in the convolution operation with the embedding matrix of \texttt{:IronMan}. Each convolution step represents the interaction between a specific feature of \texttt{:hasWife} and \texttt{:IronMan}, enabling dynamic convolution between entity and relation. ConvD integrates the dynamic convolution results based on priori knowledge and compares them with the candidate entity embedding matrices. The experimental results show that the model predicts that \texttt{Pepper Potts} is Iron Man’s wife with a probability of 83.206\%, achieving the correct prediction.

\renewcommand{\arraystretch}{1.1}
\begin{table*}[ht]
    \begin{center}
    \caption{Effect of dynamic convolution}
    \label{dynamic}
        \begin{tabular}{c|cccc|cccc|cccc}  
            \hline
            \multirow{2}{*}{\textbf{Model}} & \multicolumn{4}{c|}{\textbf{FB15K-237}} & \multicolumn{4}{c|}{\textbf{WN18RR}}& \multicolumn{4}{c}{\textbf{DB100K}} \\
            \cline{2-13}
            ~ & \textbf{MRR} & \textbf{Hits@10} & \textbf{Hits@3} & \textbf{Hits@1} & \textbf{MRR} & \textbf{Hits@10} & \textbf{Hits@3} & \textbf{Hits@1} & \textbf{MRR} & \textbf{Hits@10} & \textbf{Hits@3} & \textbf{Hits@1} \\
            \hline
            \textbf{10\%} Convolution & 0.366 & 0.547 & 0.402 & 0.276 & 0.423 & 0.493 & 0.443 & 0.398 & 0.352 & 0.531 & 0.394 & 0.264\\
            \textbf{20\%} Convolution & 0.368 & 0.569 & 0.429 & 0.297 & 0.445 & 0.533 & 0.462 & 0.432 & 0.369 & 0.554 & 0.415 & 0.277\\
            \textbf{30\%} Convolution & 0.373 & 0.568 & 0.426 & 0.305 & 0.453 & 0.545 & 0.472 & 0.443 & 0.379 & 0.567 & 0.427 & 0.283 \\
            \textbf{40\%} Convolution & 0.377 & 0.567 & 0.417 & 0.301 & 0.463 & 0.545 & 0.48 & 0.452 & 0.386 & 0.577 & 0.436 & 0.289 \\
            \textbf{50\%} Convolution & 0.376 & 0.569 & 0.429 & 0.307 & 0.467 & 0.560 & 0.482 & 0.460 & 0.395 & 0.584 & 0.444 & 0.294 \\
            \textbf{60\%} Convolution & 0.381 & 0.571 & 0.431 & 0.299 & 0.469 & 0.564 & 0.488 & 0.460 & 0.401 & 0.588 & 0.453 & 0.302\\
            \textbf{70\%} Convolution & 0.387 & 0.578 & 0.441 & 0.311 & 0.475 & 0.559 & 0.488 & 0.463 & 0.410 & 0.599 & 0.459 & 0.308 \\
            \textbf{80\%} Convolution & 0.393 & 0.581 & 0.443 & 0.307 & 0.474 & 0.563 & 0.491 & 0.461 & 0.412 & 0.598 & 0.462 & 0.314 \\
            \textbf{90\%} Convolution & 0.399 & 0.579 & 0.445 & 0.317 & 0.478 & 0.571 & 0.487 & 0.462 & \textbf{0.415} & 0.604 & 0.464 & 0.312 \\
            \hline
            \textbf{ConvD(100\%)} & \textbf{0.409} & \textbf{0.588} & \textbf{0.447} & \textbf{0.320} & \textbf{0.479} & \textbf{0.587} & \textbf{0.495} & \textbf{0.465} & \textbf{0.415} & \textbf{0.607} & \textbf{0.467} & \textbf{0.315} \\
            \hline
        \end{tabular}
    \end{center}
\end{table*}

\subsection{Comparative Analysis of ConvD and LLM}

LLMs have demonstrated significant advantages across various NLP tasks, which has led to the exploration of their application in the field of KGC.  However, the strength of these models is predominantly evident in natural language generation, while their application to classification-based tasks, such as predicting multiple candidate entities, does not fully capitalize on the potential of LLMs.

Recent studies have proposed an innovative approach wherein structured data in the form of triples is converted into query-based text. By querying LLMs regarding the logical validity of a sentence, these models can be used to assess the truthfulness of a triple. Specifically, the triple, which consists of a head entity, a relation, and a tail entity, is posed as a query in the form: “Given a triple from a knowledge graph. The triple consists of head entity \texttt{Iron Man}, relation \texttt{hasWife}, and tail entity \texttt{Pepper Potts}. Please determine the correctness of the triple and respond True or False.” The LLM’s output is then analyzed: if the response includes "True," the triple is considered valid; otherwise, it is deemed false, as shown in Figure ~\ref{fig:case}.

KoPA \cite{KoPA} introduces an enhancement to existing link prediction methods by employing metrics such as accuracy and F1 score to evaluate the performance of LLMs in the link prediction task. This approach has achieved accuracy rates exceeding 70\%. However, it is important to note that this binary classification task, in comparison to traditional link prediction ranking tasks, is considerably simplified. In this context, the model is only required to predict whether a candidate entity matches a given triple, rather than determining its rank within a set of candidate entities, thereby reducing the complexity of the task significantly.

\begin{table}[ht]
    \begin{center}
      \caption{Results of comparative experiments}
      \label{tab:comparative-experiments}  
        \begin{tabular}{c|cc|cc}
        \toprule
        \multirow{2}{*}{\textbf{Model}} & \multicolumn{2}{c|}{\textbf{FB15K-237}} & \multicolumn{2}{c}{\textbf{WN18RR}} \\
        \cline{2-5}
         & \textbf{MRR} & \textbf{Hits@1} & \textbf{MRR} & \textbf{Hits@1} \\
        \hline
        KoPA~\cite{KoPA} & 0.243 & 0.207 & 0.277 & 0.238 \\
        ConvD (Ours) & \textbf{0.409} & \textbf{0.320} & \textbf{0.479} & \textbf{0.465} \\
        \bottomrule
      \end{tabular}
    \end{center}
\end{table}

We reformulate the binary classification task of LLMs into a ranking prediction task. Specifically, we request the LLMs to rank candidate entities by providing most of the relevant entities within the query. The performance is then evaluated using metrics such as MRR and Hit@1, with the experimental results presented in Table ~\ref{tab:comparative-experiments}. The experimental results show that LLMs significantly underperform compared to current specialized link prediction models in the pure ranking prediction task. Specifically, on the FB15K-237 dataset, the performance of LLMs is only 62.05\% of that achieved by ConvD, while on the WN18RR dataset, it is 54.51\%. We hypothesize that knowledge graphs, being highly structured data, present a challenge for LLMs. These models, while trained on vast amounts of data, rely on learned features and existing knowledge matching rather than adhering to fixed logical structures, which limits their ability to predict accurate rankings in such structured tasks.

\subsection{Parameter Analysis}

\subsubsection{Dynamic Convolutional Interaction Exploration}

To validate the effectiveness of dynamic convolution strategies, we deliberately reduced the number of convolutional kernels and employed an incomplete dynamic convolution strategy for testing, aiming to assess the impact of incomplete dynamic convolutional interactions on model performance. Specifically, we constrained the number of kernels to 90\%, 80\%, etc., relative to the optimal performance and obtained corresponding performance metrics for the incomplete dynamic convolutional interaction models. This experiment was conducted on the FB15K-237 and WN18RR datasets, maintaining the same hyperparameters as in previous experiments. The experimental results indicate that incomplete dynamic convolutional interactions indeed negatively affect model performance (as shown in Table \ref{dynamic}). However, as the extent of convolutional interactions increases, the negative impact diminishes, even when the number of kernels is insufficient. On FB15K-237, performance variations among different models are minimal. In contrast, on WN18RR, performance remains stable as long as the number of kernels is above 60\%; dropping below this threshold leads to significant performance degradation. This highlights the importance of dynamic convolution kernels in enhancing feature extraction capabilities, particularly in WN18RR datasets, which have fewer relations and low relation-specific in-degrees, thereby confirming the efficacy of the dynamic convolution approach.  

\subsubsection{The Size of Dynamic Convolutional Kernels}

\begin{figure}[h]
  \centering
	\includegraphics[width=83mm]{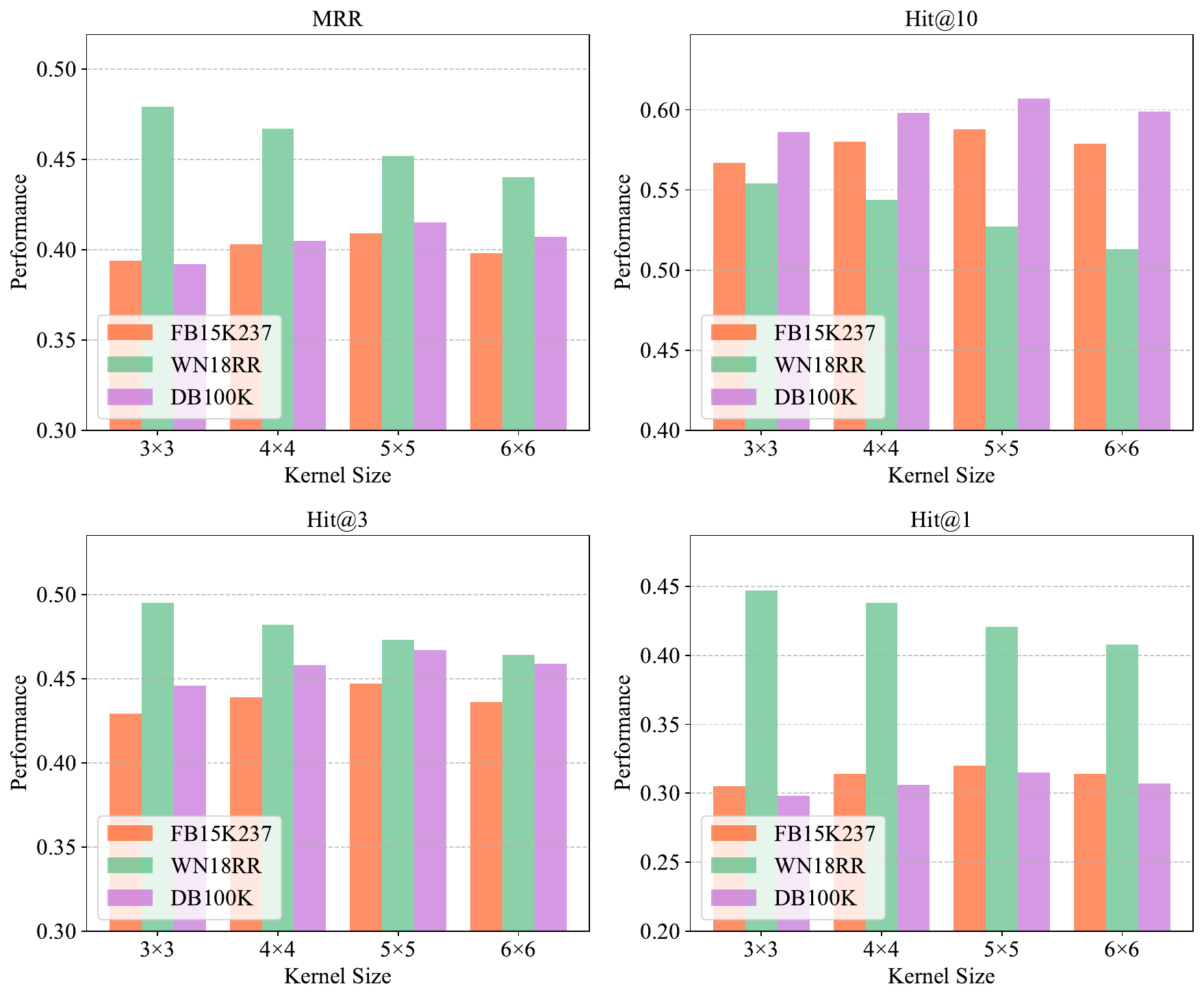}
    \caption{Results from experiments with different convolution kernel sizes.}
  \label{fig:SIZE}
\end{figure}

In this paper, we explore the effect of the number of dynamic convolution kernels on knowledge embedding performance. In addition, the size of the convolution kernel also affects the effectiveness of dynamic convolution to a certain extent, so we design parameter analysis experiments to explore the impact of dynamic convolution kernel on model performance.
We select three standard knowledge graph datasets, FB15K-237, WN18RR and DB100K, and choose the same hyperparameters as the link prediction task to perform experiments on the impact of dynamic convolution kernel sizes, and the results of the experiments are presented in Figure~\ref{fig:SIZE}.
The model performs optimally with different convolution kernel sizes for FB15K-237 (5×5), WN18RR (3×3) and DB100K(5×5) datasets, when other experimental conditions are held constant. This difference reflects the reliance on relation embeddings’ dimensionality, especially in datasets with fewer relations. Increasing the number and size of dynamic convolution kernels enhances feature extraction in datasets with fewer relations. Conversely, in datasets with a larger number of relations, smaller kernels suffice as larger ones may introduce noise, reducing the model's effectiveness.

\begin{figure}[h]
  \centering
	\includegraphics[width=83mm]{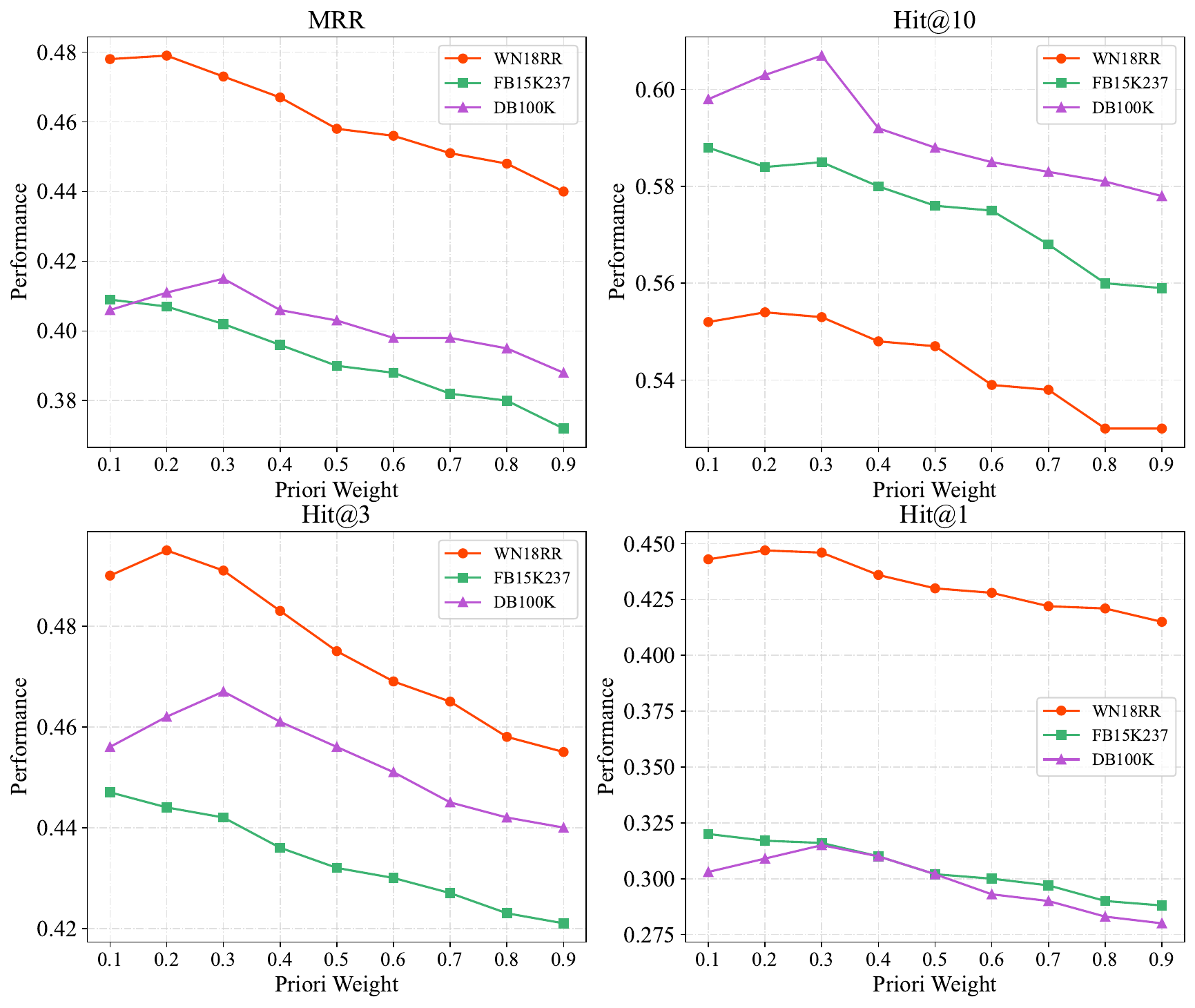}
    \caption{Results from experiments with different priori knowledge weights.}
  \label{fig:P}
\end{figure}

\begin{figure}[h]
  \centering
	\includegraphics[width=83mm]{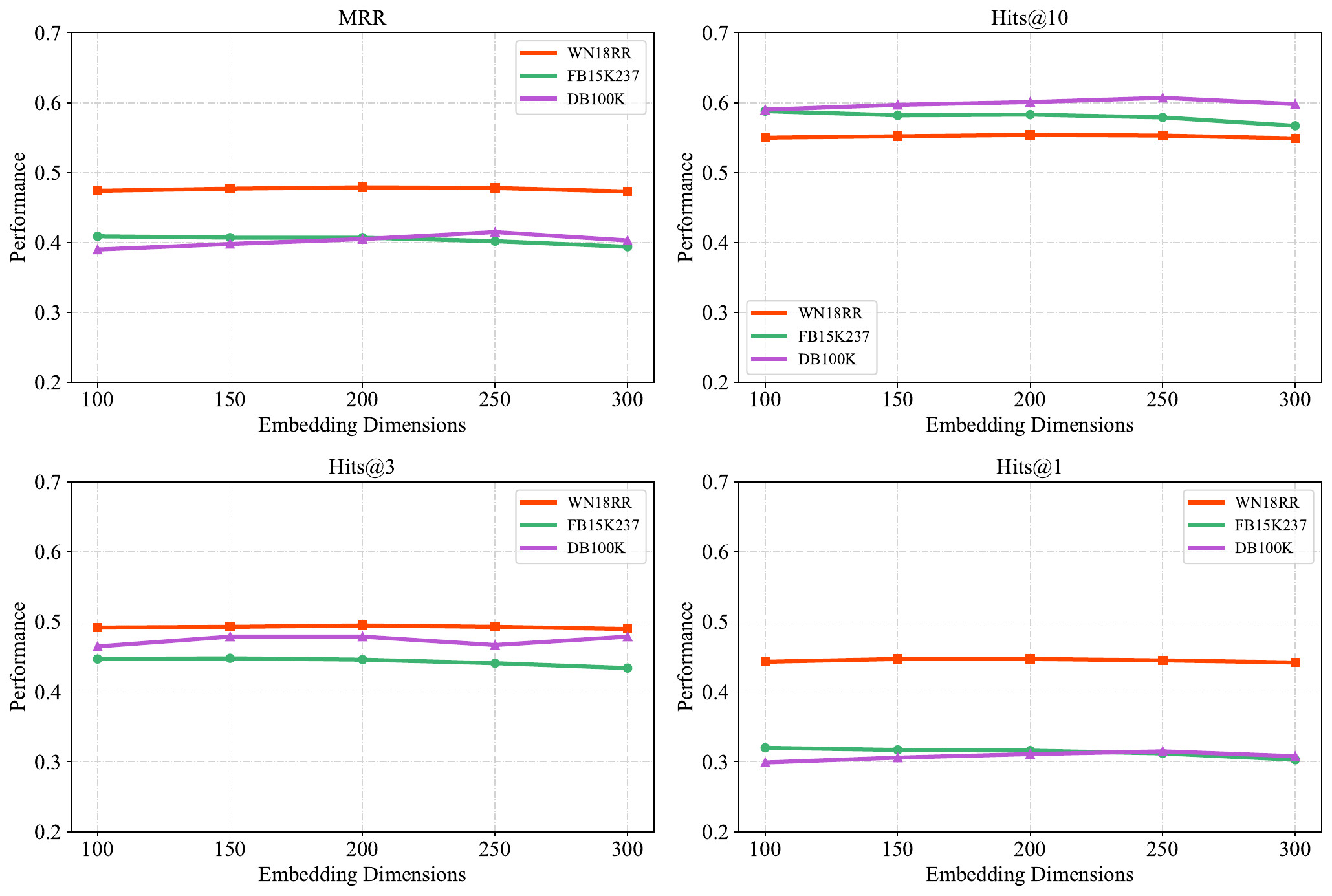}
    \caption{Results of experiments performed with different dimensions of entity embedding.}
  \label{fig:WEI}
\end{figure}

\subsubsection{Priori Knowledge and Embedding Dimension}

Meanwhile, we will explore the influence of a priori knowledge and embedding dimension on the model effectiveness. On three standard knowledge graph datasets, we conduct experiments by selecting the same hyperparameters as the link prediction experiments with a priori knowledge embedding weights and entity embedding dimensions as variables respectively, and the experimental results are displayed in Fig \ref{fig:P} and Fig \ref{fig:WEI}.
In KGC models, moderately incorporating priori knowledge of entity-relation pairs can effectively enhance model performance. However, assigning excessive weight to this priori knowledge may introduce noise and consequently degrade performance. The optimal weight for priori knowledge varies across different datasets. For instance, FB15K-237 and DB100K achieve the best performance with a relatively low weight (approximately 0.1), while WN18RR prefers a slightly higher weight (around 0.2).
The experiments on the embedding dimension indicate that the optimal performance of ConvD is achieved at a dimension of 100 on the FB15K-237 dataset, 200 on the WN18RR dataset, and 250 on the DB100K dataset, and the model effectiveness shows a certain degree of decline with the change of increasing and decreasing entity embedding dimensions. Since the number of relations in the WN18RR dataset is far less than that in the FB15K-237 dataset, more dimensions are required in order to fully interact with the restricted number of relation-aware dynamic convolution kernels and extract more feature.

\section{Conclusion}
In this paper, we propose a dynamic convolutional embedding method ConvD for knowledge graph completion, which mainly constructs multiple relation-aware internal convolution kernels to increase feature interactions. Additionally, ConvD uses an attention mechanism optimized with priori knowledge to assign different weight coefficients to multiple relation convolution kernels, improving the model's generalization ability. Experiments on different datasets show that our proposed model has consistent and significant superiority and robustness over other state-of-the-art methods.

\bibliographystyle{IEEEtran}
\bibliography{reference}


\begin{IEEEbiography}[{\includegraphics[width=1in,height=1.25in,keepaspectratio]{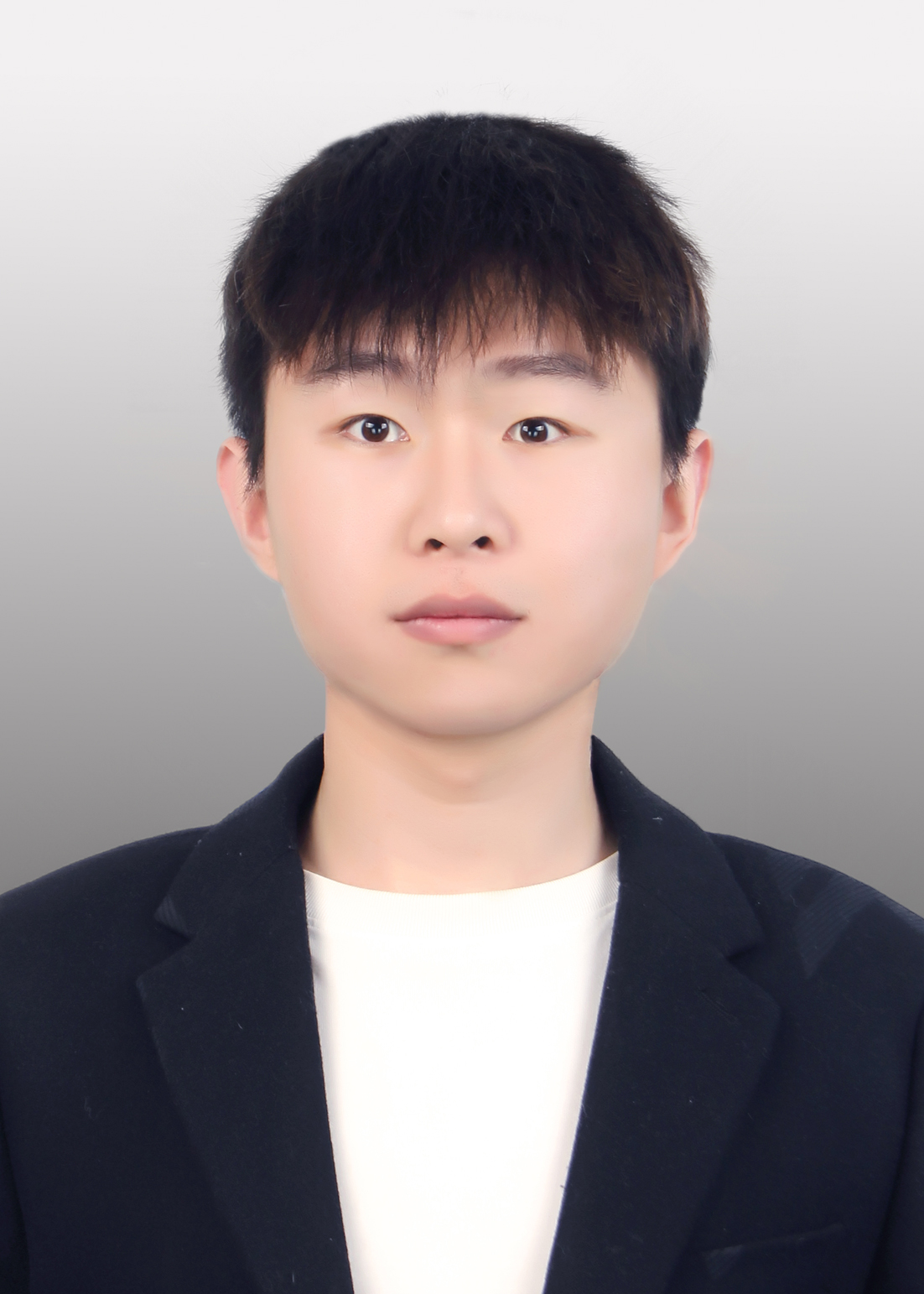}}]{Wenbin Guo}
is currently pursuing a PhD degree in Computer Science and Technology, College of Intelligence and Computing, Tianjin University, Tianjin, China. His current research interests include knowledge representation learning.
\end{IEEEbiography}

\begin{IEEEbiography}[{\includegraphics[width=1in,height=1.25in,keepaspectratio]{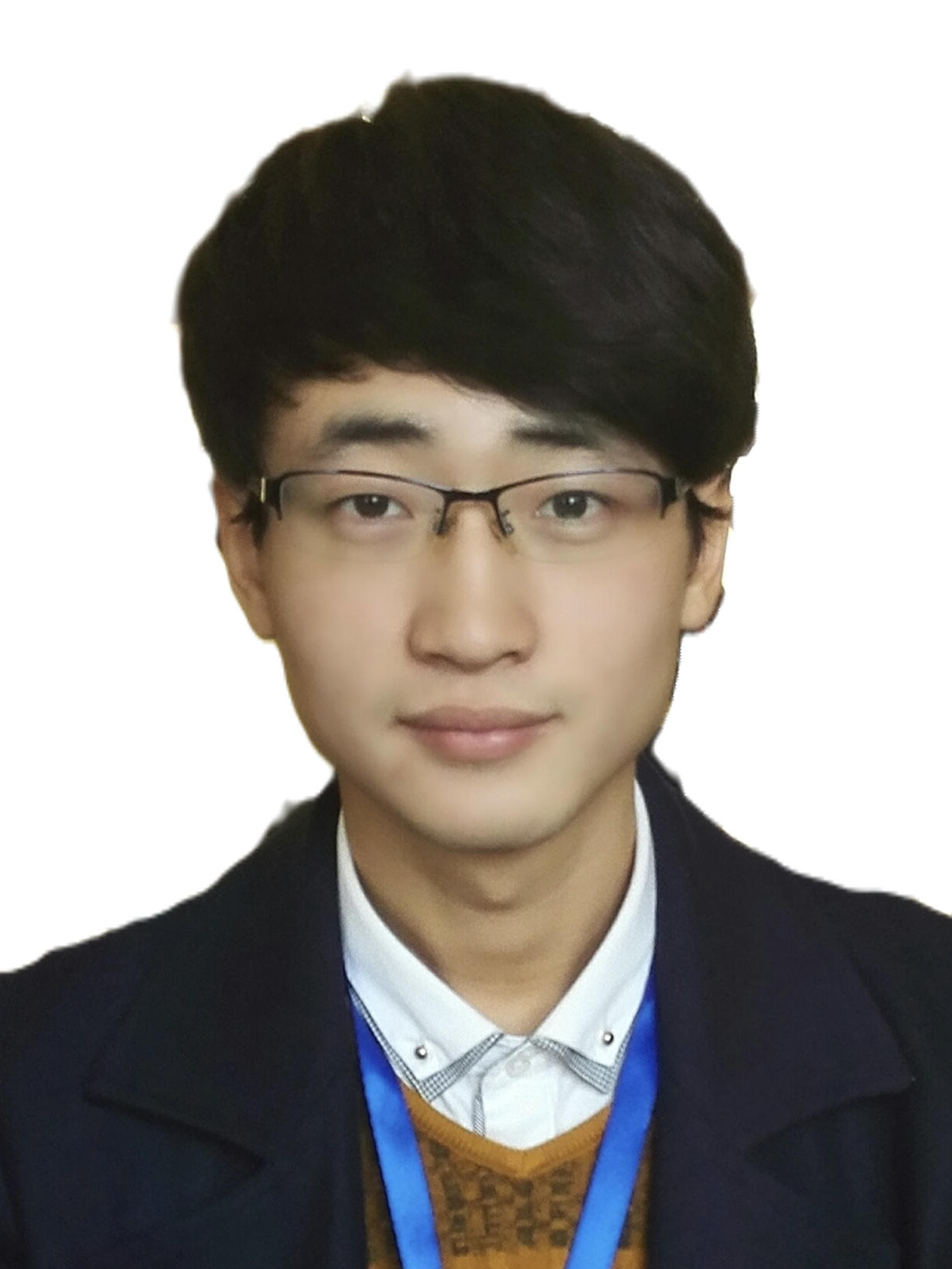}}]{Zhao Li}
is currently pursuing a PhD degree in Computer Science and Technology, College of Intelligence and Computing, Tianjin University, Tianjin, China. His current research interests include knowledge graph, representation learning and reasoning, and Responsible AI. He is the president of the Tianjin University Student Chapter of the China Computer Federation.
\end{IEEEbiography}

\begin{IEEEbiography}[{\includegraphics[width=1in,height=1.25in,keepaspectratio]{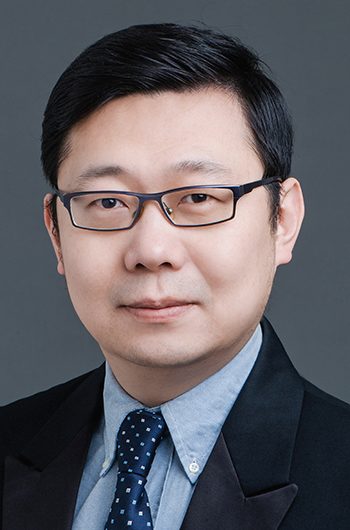}}]{Xin Wang}
received the BE and PhD degrees in computer science and technology from Nankai University in 2004 and 2009, respectively. He is currently a professor in the College of Intelligence and Computing, Tianjin University. His research interests include knowledge graph data management, graph databases, and big data distributed processing. He is a member of the IEEE and the IEEE Computer Society.
\end{IEEEbiography}

\begin{IEEEbiography}[{\includegraphics[width=1in,height=1.25in,keepaspectratio]{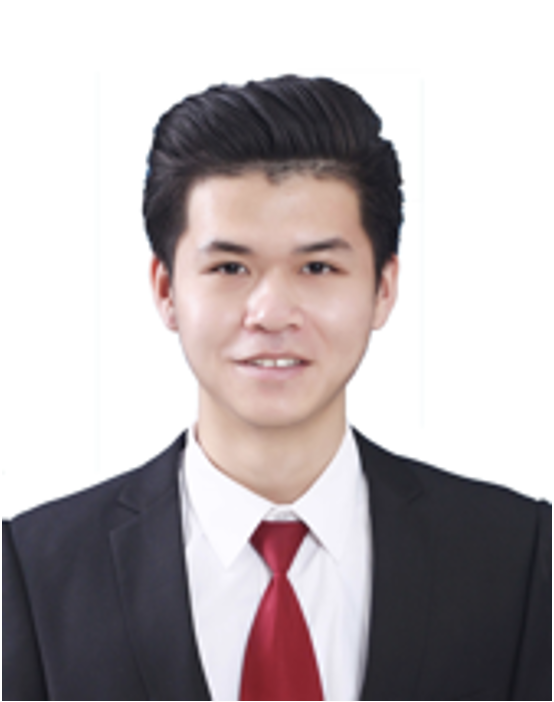}}]{Zirui Chen}
is currently pursuing a PhD degree in Computer Science and Technology, College of Intelligence and Computing, Tianjin University, Tianjin, China. His current research interests include knowledge representation learning and knowledge graph question answering.
\end{IEEEbiography}

\begin{IEEEbiography}[{\includegraphics[width=1in,height=1.25in,keepaspectratio]{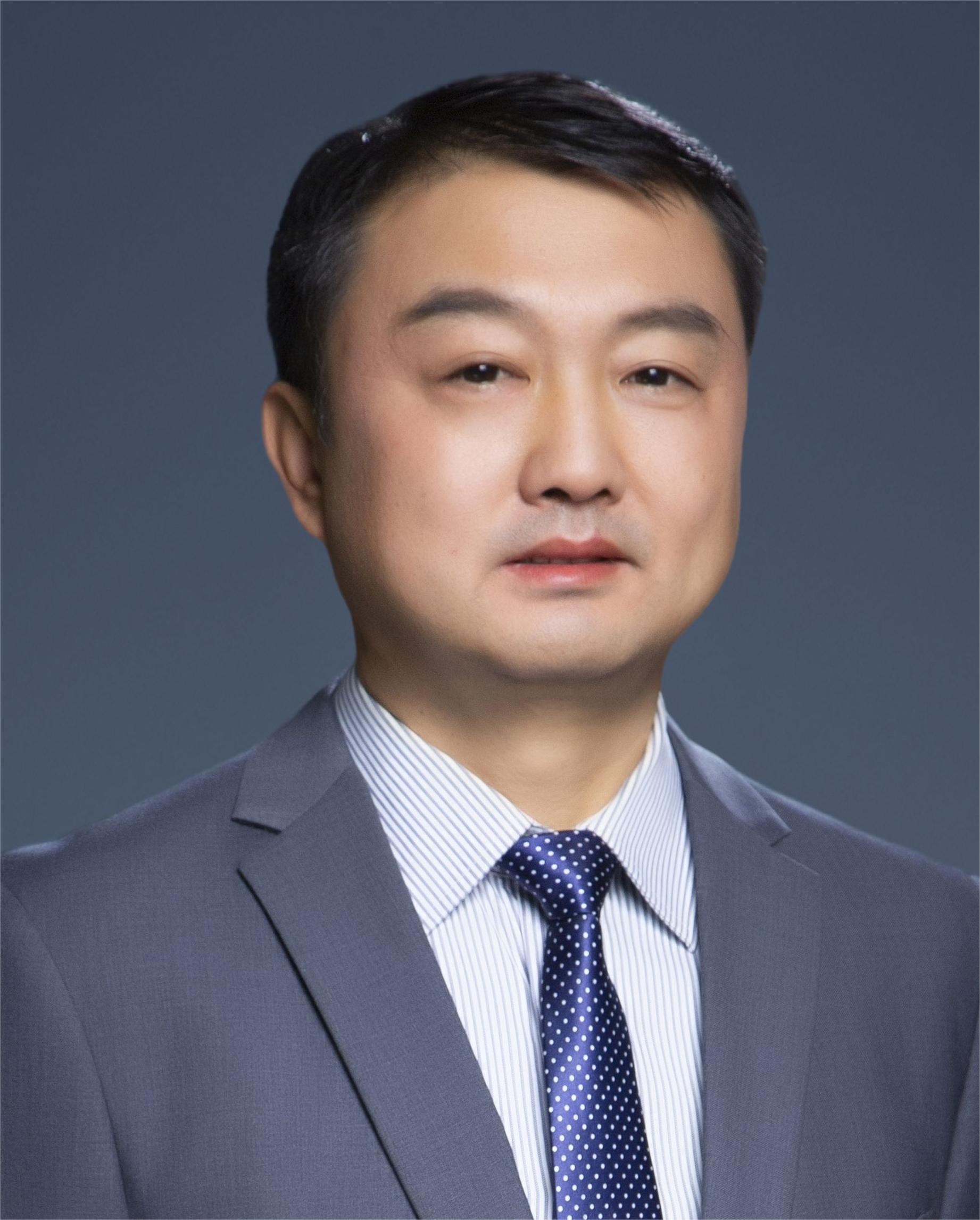}}]{Jun Zhao}
received Ph.D. degree in management science and engineering from the Beijing Institute of Technology, China, in 2006. He is the former dean of the business school and currently is a professor in the economic and management of school, Ningxia University. His research interests include information system, computational experiments, and big data intelligent applications.
\end{IEEEbiography}

\begin{IEEEbiography}[{\includegraphics[width=1in,height=1.25in,keepaspectratio]{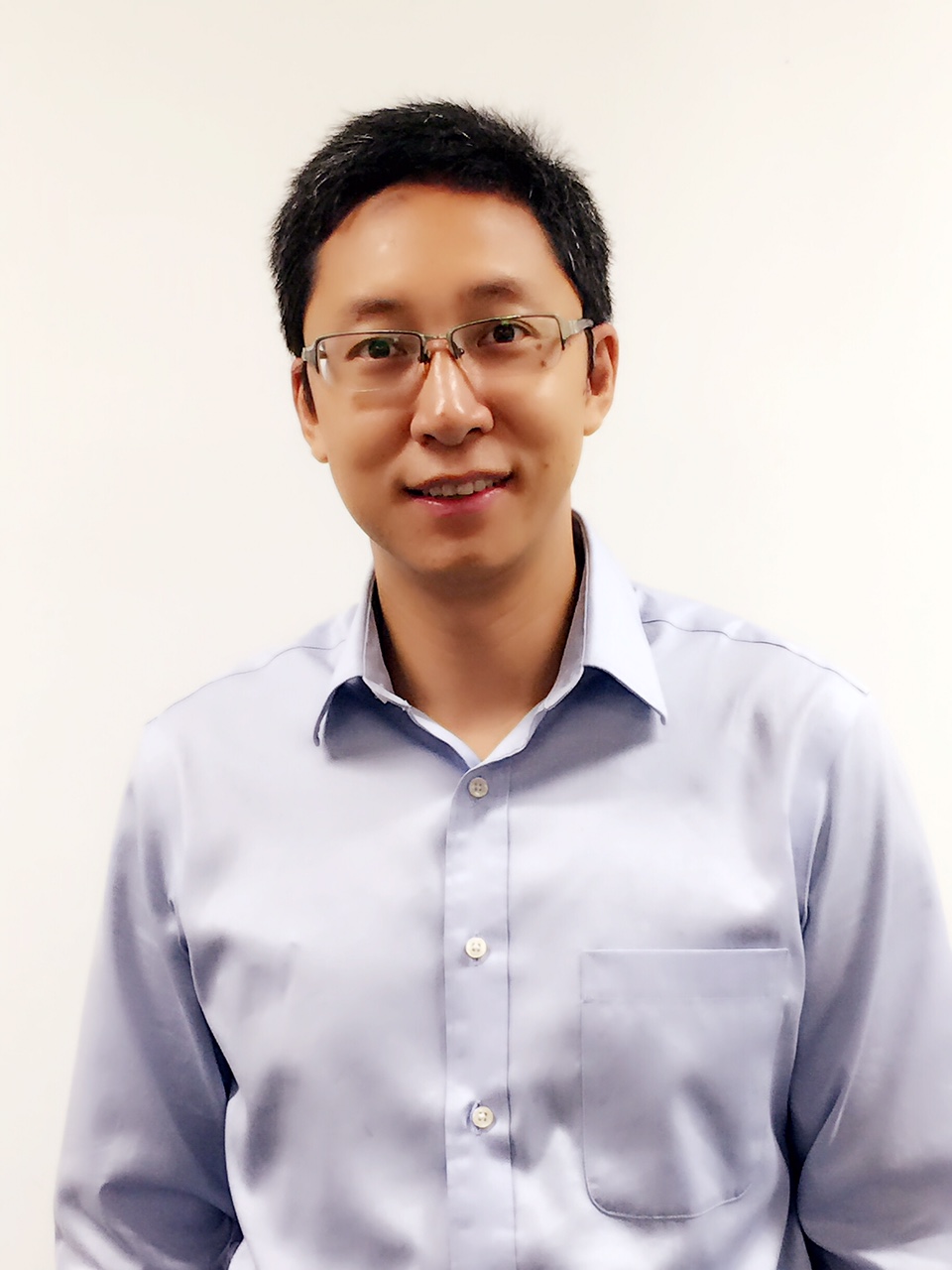}}]{Jianxin Li}
received a Ph.D. degree in computer science from the Swinburne University of Technology, Melbourne, VIC, Australia, in 2009. He is a Professor in the Discipline of Business Systems and Operations, Edith Cowan University, Joondalup, Australia. He was awarded as World Top 2\% scientists by 2023 Stanford due to his research impact and high citation. His current research interests include database query processing and optimization, social network analytics, and traffic network data processing. He is a senior member of the IEEE Computer Society.
\end{IEEEbiography}

\begin{IEEEbiography}[{\includegraphics[width=1in,height=1.25in,keepaspectratio]{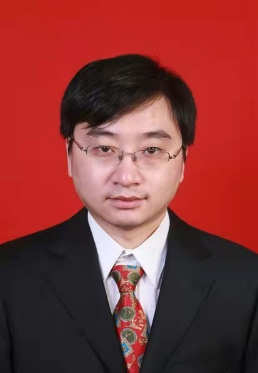}}]{Ye Yuan}
received the BS, MS, and PhD degrees in computer science from Northeastern University, in 2004, 2007, and 2011, respectively. He is currently a professor with the Department of Computer Science, Beijing Institute of Technology, China. His research interests include graph embedding, graph neural networks, and social network analysis. He won the National Science Fund for Excellent Young Scholars, in 2016.


\end{IEEEbiography}

\vfill

\end{document}